\documentclass[a4paper,fleqn]{cas-dc}

\usepackage[numbers]{natbib}
\usepackage{amsmath,amssymb,amsfonts}
\usepackage{algorithmic}
\usepackage{graphicx}
\usepackage{subcaption}
\usepackage{textcomp}
\usepackage{xcolor}
\usepackage{url}
\usepackage{booktabs}
\usepackage{multirow}
\usepackage{algorithm}
\usepackage{hyperref}
\usepackage{comment}
\usepackage{tabularx}
\usepackage{bm}

\usepackage{placeins}   

\begin{document}
\let\WriteBookmarks\relax

\shorttitle{Learning Social Robot Navigation By Sensing Human Legs}    
\shortauthors{A. Vaglio et al.}  

\title[mode = title]{Learning Social Robot Navigation By Sensing Human Legs}  

\author[1]{Alberto Vaglio}[orcid=0009-0000-8718-2211]
\cormark[1]
\ead{alberto.vaglio@student.unisi.it}
\credit{Conceptualization, Methodology, Software, Validation,
        Formal analysis, Investigation, Data curation, Visualization,
        Writing -- original draft}

\author[1]{Andrea Garulli}
\credit{Supervision, Writing -- review \& editing. Validation, Supervision, Methodology, Investigation, Formal analysis, Conceptualization}

\author[1]{Antonio Giannitrapani}
\credit{Supervision, Writing -- review \& editing. Validation, Supervision, Methodology, Investigation, Formal analysis, Conceptualization}

\author[2]{Renato Quartullo}
\credit{Supervision, Writing -- review \& editing. Validation, Supervision, Methodology, Investigation, Formal analysis, Conceptualization}

\author[1]{Tommaso Van Der Meer}
\credit{Supervision, Writing -- review \& editing. Validation, Supervision, Methodology, Investigation, Formal analysis, Conceptualization}

\affiliation[1]{organization={Department of Information Engineering and Mathematics},
            addressline={University of Siena},
            city={Siena},
            country={Italy}}

\affiliation[2]{organization={Uninettuno University},
            city={Rome},
            country={Italy}}

\cortext[1]{Corresponding author}

\begin{abstract}
Robots navigating among pedestrians typically sense their surroundings with a 2D LiDAR mounted close to the ground. At that height, the sensor mostly sees moving legs rather than whole people, yet most learning-based navigation methods still treat pedestrians as simple shapes like circles. This paper addresses that gap with CALF (Convolutional Attention for Leg Features), an end-to-end neural architecture that combines convolutional layers, attention, and MLP to interpret leg motion directly from LiDAR scans and produce safe navigation commands. The CALF policy is trained using deep reinforcement learning algorithms within LegNav, a custom lightweight 2D simulator that combines 2D LiDAR ray tracing with a novel pedestrian gait model. The resulting policy is compared against classical and learning-based baselines in terms of navigation performance and social compliance. The approach is validated through real-world experiments via zero-shot deployment on a TurtleBot 4, yielding smooth and socially compliant trajectories. Written in JAX, the LegNav simulator enables the training of a deployment-ready CALF policy in under an hour on a single consumer GPU.
\end{abstract}


\maketitle

\section{Introduction}

Social navigation remains one of the fundamental challenges for mobile robots operating in populated environments. 
\begin{figure}[pos=!tp]
     \centering
     \begin{subfigure}[b]{\columnwidth}
         \centering
         \includegraphics[width=0.88\columnwidth,height=0.28\textheight,keepaspectratio]{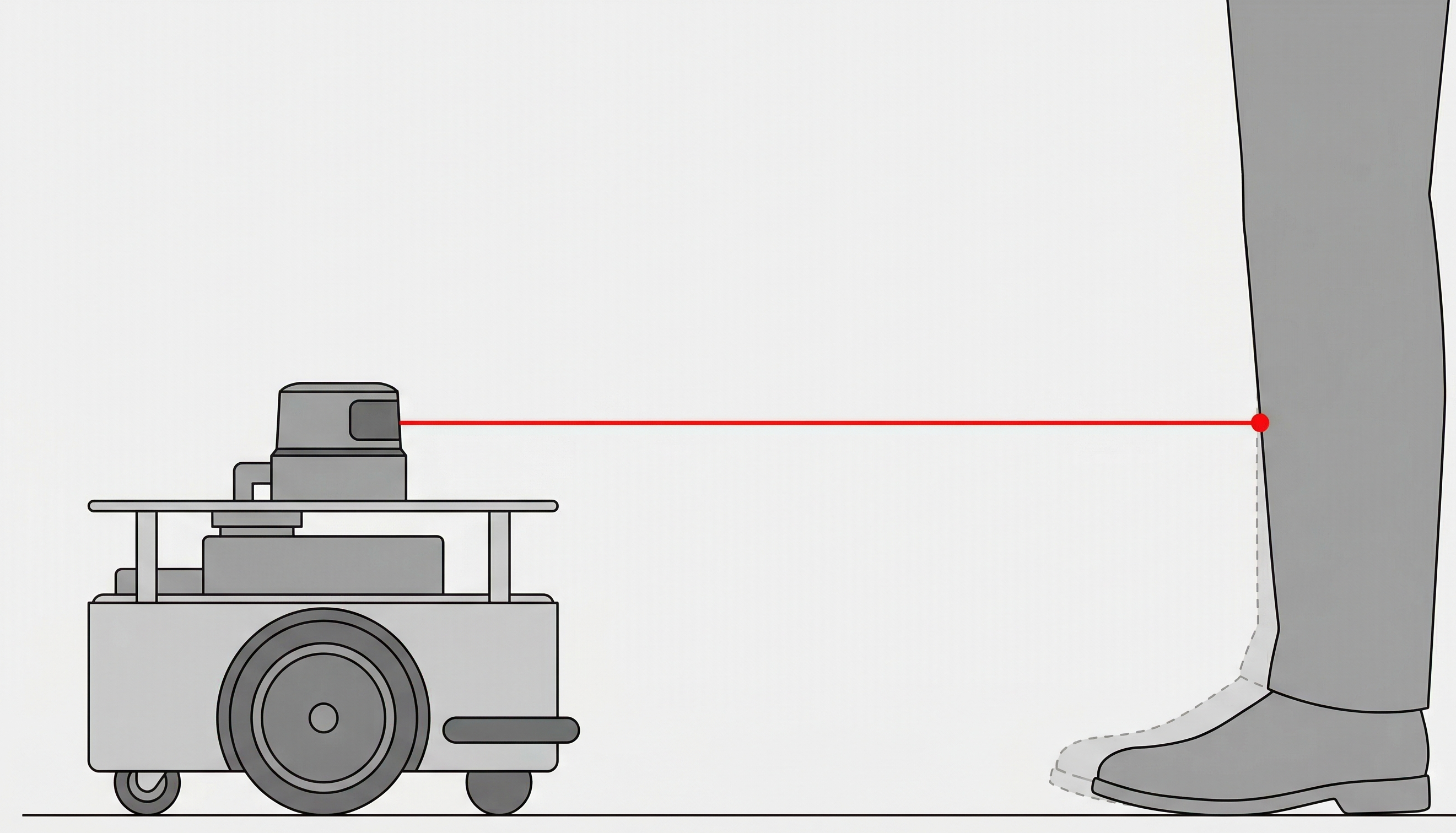}
         \caption{}
         \label{fig:lidar_leg_occlusion}
     \end{subfigure}
     \vspace{0.5em}
     \begin{subfigure}[b]{\columnwidth}
         \centering
         \includegraphics[width=0.88\columnwidth,height=0.28\textheight,keepaspectratio]{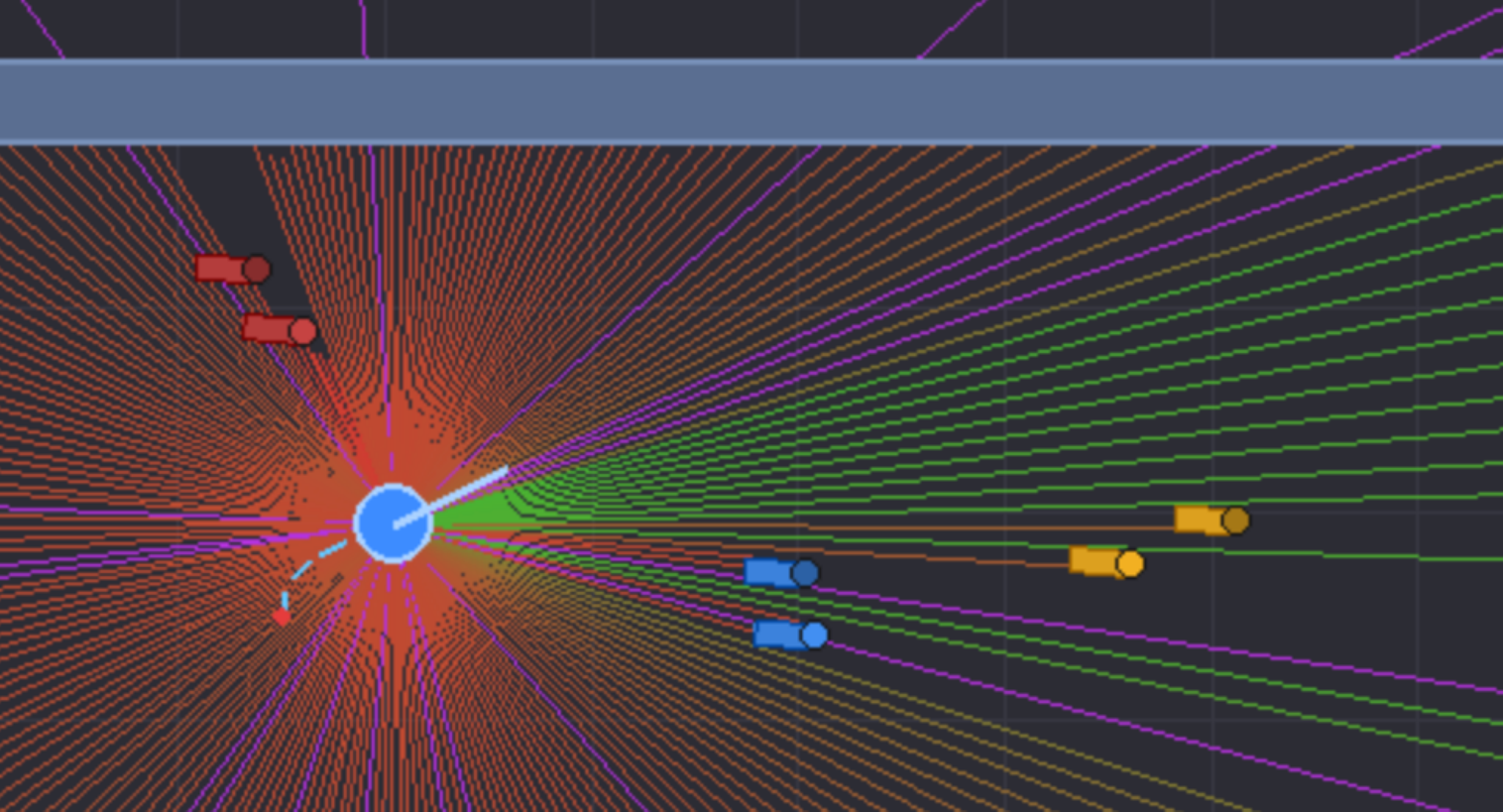}
         \caption{}
         \label{fig:sim_env}
     \end{subfigure}
     \caption{(a) Side view of a mobile robot equipped with an ankle-height LiDAR. The beam intercepts the nearest boundary of the pedestrian's leg, while the shoe, protruding below the scan plane, remains undetected. (b) The LegNav simulation environment, featuring explicit human leg dynamics and realistic LiDAR returns from individual leg pairs: the two leg footprints of each pedestrian are separately visible in the LiDAR sweep.}
     \label{fig:total_figure}
\end{figure}
Unlike static obstacle avoidance, social navigation requires handling human motion, respecting interpersonal space and reacting safely to unpredictable behaviors. Classical navigation pipelines, combining global planners with local reactive methods, provide strong guarantees in structured environments. However, they struggle with dense human traffic as they lack the ability to deal with dynamic obstacles.

This challenge increases significantly when relying only on 2D LiDAR data.
Planar scans miss the semantic richness needed to distinguish obstacles; therefore, detecting pedestrians becomes difficult when they appear merely as clusters of range returns. Furthermore, the vast majority of mobile robots are equipped with 2D LiDAR sensors mounted between 10 and 20~cm above the ground \cite{arras2007using, leigh2015person}. At this scanning height, the horizontal laser plane intersects only the lower legs of nearby pedestrians: most beams either hit one of the two shins or pass through the gap between them. The robot never perceives the pedestrian's torso, arms, or full body outline, but only two small, independently moving point clusters, one per leg \cite{arras2007using}. This partial observability introduces a critical blind spot at close range. As Fig.~\ref{fig:lidar_leg_occlusion} illustrates, the LiDAR intercepts the leg cylinder but misses the shoe protruding below the scan plane: a standard adult shoe extends 20--30~cm forward from the ankle, creating a significant, undetected obstacle during close-range interactions. Despite this, most social navigation frameworks approximate humans as simple circular agents, completely ignoring leg dynamics and foot geometry.

To capture these two effects at the sensor level, we propose the Non-Slip Gait (NSG) model, which represents each pedestrian as two alternatively moving feet rather than a single disc. As the body advances, the two feet alternate between a stance and a swing phase, at a variable stepping rate. Each foot is assigned its own orientation and a rectangular footprint that defines the shoe geometry and reproduces the protruding zone sitting below the scan plane, invisible to the LiDAR. The model thus recreates the leg-level structure and the foot occlusion that disc-based frameworks ignore. 
The NSG model is embedded in LegNav, a custom lightweight 2D simulation environment, in which each foot is ray traced individually by the simulated 2D LiDAR (Fig.~\ref{fig:sim_env}).

To effectively handle the complex ankle-height LiDAR observations
generated by pedestrians, end-to-end navigation policies can be employed
to directly map raw sensor measurements to velocity commands. Within a
Deep Reinforcement Learning (DRL) framework, these policies are
typically represented by neural networks with various architectures that
are trained through interactions with a simulator of the environment.
Such networks learn to implicitly extract relevant features from the
data, bypassing the need for explicit pedestrian detection, tracking, or
trajectory prediction modules. In this work, we propose CALF
\emph{(Convolutional Attention for Leg Features)}, a novel hybrid
end-to-end architecture that combines convolutional neural networks
(CNNs), self-attention mechanisms, and a multilayer perceptron (MLP). In
this network, a weight-shared 1D convolutional encoder processes a stack
of consecutive LiDAR frames, extracting leg-related features. This is
followed by a temporal self-attention module that implicitly infers leg
motion across frames without explicit tracking.

Among the DRL approaches that can be used to train networks,
Actor-Critic methods have shown promising results in robotics, and in
particular in social robot navigation. On-policy methods such as
Proximal Policy Optimization (PPO) \cite{schulman2017proximal} provide stable training, while
off-policy methods such as Soft Actor-Critic (SAC) \cite{haarnoja2018soft} improve sample
efficiency. More recently, distributional RL approaches such as
Truncated Quantile Critics (TQC) \cite{kuznetsov2020controlling} have demonstrated increased
robustness in stochastic control tasks by modeling return distributions.
All such methods are considered in the paper for training the CALF network.

Much of the existing literature focuses on training navigation policies without explicitly teaching the agent to adopt social norms. These behaviors are instead left to emerge naturally through the learning process \cite{chen2017decentralized, chen2019crowd, everett2018motion}. In this work instead, we explicitly enforce the adoption of social conventions, specifically \textit{yielding} (i.e., the robot stopping and waiting when a person is in close proximity), by directly embedding this behavior into the RL agent reward. Furthermore, we introduce the Yielding Score as a metric for evaluating the adherence of the trained policy to the desired behavior.

Summing up, the contribution of this paper is threefold:
(i) we propose CALF, a novel hybrid end-to-end architecture that blends 1D-CNNs, attention mechanisms, and an MLP, designed specifically for social robot navigation with human leg dynamics, based on 2D LiDAR data;
(ii) we propose the NSG model for simulating human gait dynamics in 2D environments;
(iii) we introduce LegNav, a custom 2D social navigation simulator with explicit human leg dynamics and 2D LiDAR ray tracing, capable of running thousands of parallel environments and training competitive policies in less than one hour on a single NVIDIA consumer GPU, thanks to a JAX-based implementation \cite{jax2018github}.

Simulation results show that the end-to-end CALF policies perform well compared to both RL-based and classical social navigation approaches, through a benchmark evaluating navigation efficiency, safety, kinematic smoothness and social compliance.
To further validate the trained policies we zero-shot deployed the best one on a real TurtleBot~4 platform, demonstrating socially compliant behavior in human-populated indoor environments.

The remainder of this paper is organized as follows. Section~\ref{sec:related} reviews the related literature on social robot navigation. Section~\ref{sec:problem} formulates the navigation problem and defines the observation and action spaces. Section~\ref{sec:env} describes the simulation environment in detail, including the human dynamics model and the NSG model. Section~\ref{sec:navigation_policy} presents the CALF network architecture, the DRL algorithms and the reward shaping. Section~\ref{sec:training} explains the training infrastructure. Section~\ref{sec:experiments} reports experimental results, and Section~\ref{sec:conclusion} concludes the paper.

The complete open-source codebase, including the trained policies, is publicly available\footnote{\url{https://github.com/otr-ebla/LegNav-Sim}}.
\section{Related Work}
\label{sec:related}

In this section we categorize prior work on social robot navigation into Model-based approaches, Human-Aware RL approaches, and end-to-end RL approaches.

\subsection{Model-based Approaches}
\label{sec:non-rl-based-approaches}
Classical approaches to social navigation generally fall into two distinct paradigms: mathematical models of human interaction, and robotic local planners.
The first category comprises interaction models designed to describe and simulate human crowd behavior. The Social Force Model (SFM) \cite{helbing1995social} and the Headed Social Force Model (HSFM) \cite{farina2017walking} describe pedestrian motion through a superposition of attractive goal forces and repulsive inter-agent forces. Similarly, Velocity Obstacle (VO) methods \cite{van2011reciprocal} and their reciprocal variants (RVO/ORCA) compute collision-free velocities by reasoning about the relative velocity space of all agents. Because their primary function is modeling human dynamics, they are widely adopted as the simulation backbone for navigation benchmarks. While these mathematical models can be adapted to act as a navigation controller for an autonomous robot, they cannot ingest raw sensor data directly. Instead, they require explicit tracking pipelines to estimate the precise positions and velocity states of surrounding pedestrians, alongside the positions of static obstacles. The second category consists of algorithms strictly designed for autonomous robot control. The Dynamic Window Approach (DWA) \cite{fox1997dynamic} is a standard local planner that samples feasible velocity commands within the robot's kinematic constraints, selecting the command that best balances goal-seeking and obstacle avoidance. Model Predictive Path Integral (MPPI) \cite{williams2017model} is a prominent sampling-based variant of Model Predictive Control (MPC) that excels at handling general non-linear dynamics and non-convex cost functions. Unlike interaction models, DWA and MPPI can operate directly on raw or semi-raw sensor inputs (such as LiDAR point clouds or local costmaps) for real-time obstacle avoidance. However, because they treat all range returns equally as generic obstacles, they lack intrinsic mechanisms for social compliance.

Therefore, a fundamental gap remains: the methods capable of social modeling (SFM/ORCA) rely on explicit pedestrian tracking, while the planners capable of directly processing sensor streams (DWA/MPPI) remain socially blind and furthermore require hand tuning of heuristic parameters. This highlights the need for an approach that can map raw sensor observations directly to socially compliant navigation commands.

\subsection{Human-Aware RL Approaches}
Human-Aware Reinforcement Learning methods tackle social navigation by explicitly incorporating the kinematic states of surrounding humans, such as their positions and velocities, directly into the robot's state representation. Early approaches in this domain utilized discrete action spaces and handcrafted state features, assuming full observability of the crowd dynamics. For example, CADRL \cite{chen2017decentralized} formulates pairwise collision avoidance as a sequential decision problem and trains a value network using self-play. SA-CADRL \cite{chen2017socially} extends this approach to multi-agent settings by introducing a socially aware value function. SARL \cite{chen2019crowd} further incorporates attention mechanisms to weight the influence of individual pedestrians on the navigation policy. GA3C-CADRL \cite{everett2018motion} scales these approaches to larger crowds using GPU-accelerated asynchronous advantage actor-critic training, while LSTM-based crowd models \cite{everett2021collision} introduce temporal recurrence to capture pedestrian motion history. 
Building on these foundations, \cite{chen2020relational} proposes a relational graph learning approach, which uses a graph convolutional network to model higher-order agent interactions and performs multi-step lookahead planning by predicting human trajectories, achieving notable gains in efficiency and collision reduction. Similarly, \cite{yang2023st} employs the Spatial-Temporal State Transformer ($ST^2$) to efficiently encode both global spatial interactions and temporal correlations among consecutive frames for crowd-aware navigation. More recently, NaviSTAR \cite{wang2023navistar} leverages a hybrid spatio-temporal graph transformer combined with preference learning to align navigation policies with human social norms, demonstrating strong performance both in simulation and real-world experiments. Bridging the gap between model-based representations and DRL, \cite{martinez2025rumor} introduces RUMOR, an agent-based planner that abstracts the dynamic environment into a robocentric Dynamic Object Velocity Space (DOVS).

However, a major limitation of these approaches is their reliance on full state information. Providing explicit pedestrian trajectories to the RL agent requires sophisticated, external perception and tracking systems, which are computationally expensive and prone to failure in real-world physical deployments.

\subsection{End-to-End RL Approaches}

End-to-end deep reinforcement learning methods map raw sensor data straight to robot actions. This paradigm eliminates the intermediate detection pipeline, reduces system complexity, and facilitates immediate physical deployment.

Numerous recent works address 2D LiDAR navigation by directly processing raw inputs. For example, policies have been developed to navigate tight indoor spaces \cite{perez2021robot}, coordinate multiple robots \cite{long2018towards}, and maintain personal and social safety around pedestrians \cite{jin2020mapless}. Aligning with this raw data philosophy, NavRep \cite{dugas2021navrep} trains unsupervised representation models to encode sensor observations into compact states, improving overall learning efficiency.
To handle dynamic environments, recent frameworks process sequential observations to infer motion trends. Zhu et al. \cite{zhu2024learn} employ an LSTM network to accumulate historical motion features, while \cite{de2024spatiotemporal} utilize a spatio-temporal attention module to extract scene dynamics directly from consecutive scans. Other approaches, such as the one presented in \cite{xie2023drl}, expand the sensory scope by combining camera images with LiDAR data and velocity obstacle rewards.

A common limitation of the above approaches is that they model pedestrians as featureless discs. This simplification creates a mismatch with physical reality, as robots equipped with 2D LiDAR perceive humans as alternating pairs of leg clusters rather than unified cylinders. Overcoming this discrepancy between idealized simulation modeling and actual physical sensor geometry is the primary motivation of this work.

\section{Problem Formulation}
\label{sec:problem}
In this section we describe the robot action and observation spaces and formulate the navigation task.

\subsection{Action Space}
We model the robot as a differential-drive platform represented by a disc of radius $r$.  
At time step $t$ its pose is given by position $(x_t, y_t)$ and heading angle $\theta_t$.
The control input at time step $t$ is
$\mathbf{a}_t = [ v_t, \omega_t ]^\top$,
where $v_t$ and $\omega_t$ denote respectively the commanded linear and angular velocities.  
The linear velocity is bounded in the interval $[0,\,v_{\max}]$ m/s. The angular velocity is constrained to $[-\omega_{\max},\,\omega_{\max}]$ rad/s.
The action space is therefore
\begin{equation}
\mathcal{A} = \{ (v, \omega) \in \mathbb{R}^2 \mid v \in [0,v_{\max}], \, \omega \in [-\omega_{\max},\omega_{\max}] \}.
\label{eq:action_space}
\end{equation} 
The robot pose evolves according to the discrete-time unicycle model \cite{thrun2005probabilistic}:
\begin{align}
    x_{t+1} &= x_t + \frac{v_t}{\omega_t}\left[\sin(\theta_t + \omega_t \Delta t) - \sin(\theta_t)\right], \label{eq:robot1}\\
    y_{t+1} &= y_t + \frac{v_t}{\omega_t}\left[\cos(\theta_t) - \cos(\theta_t + \omega_t \Delta t)\right], \label{eq:robot2}\\
    \theta_{t+1} &= \theta_t + \omega_t \Delta t, \label{eq:robot3}
\end{align}
with simulation step $\Delta t$.

\subsection{Observation Space}
\label{subsec:obs_space}

The observation vector $\mathbf{o}_t$ at time $t$ is divided into three main components:
\begin{equation}
\mathbf{o}_t = \left[ \underbrace{\mathbf{p}_{t}, \cdots, \mathbf{p}_{t-N+1}}_{\text{goal stack}\,},\; \underbrace{\mathbf{k}_t}_{\text{kin vec}\,},\; \underbrace{\mathbf{l}_t, \cdots, \mathbf{l}_{t-N+1}}_{\text{LiDAR stack}\,} \right]
\label{eq:observations}
\end{equation}
The goal vector $\mathbf{p}_t = [g_t^x, g_t^y, \rho_t] \in \mathbb{R}^3$ encodes the goal coordinates in the robot’s ego-frame and the goal alignment angle.
The kinematic vector $\mathbf{k}_t \in \mathbb{R}^5$ is the current kinematic state; it contains: the linear velocity $v_t$, the angular velocity $\omega_t$, the maximum linear velocity $v_{\max}$, the goal distance $d_t$ and the robot's current heading with respect to the global reference frame $\theta_t$. The maximum linear velocity $v_{\max}$ is included to design a policy which is agnostic with respect to the robot speed limits.
 Each LiDAR scan $\mathbf{l}_t$ is given by:
\[
    \mathbf{l}_t = [l_t^{(1)}, \cdots, l_t^{(L)}] \in \mathbb{R}^L,
\]
where $L$ is the number of rays per scan and $l_t^{(i)}$  is the single ray range measurement.

Temporal stacking of $N$ frames is applied to both goal vector $\mathbf{p}_j$ and LiDAR scans $\mathbf{l}_j$. The main reason behind this is to allow the policy to implicitly infer pedestrian velocities without explicit tracking. The stack is updated as a FIFO queue: at each timestep the oldest frame is discarded and the new measurement is appended.

The objective of the work is to learn a stochastic policy $\pi$ that maps observations to a probability distribution over the action space,
$\pi(\mathbf{a}_t|\mathbf{o}_t).$
The policy samples navigation commands from this distribution based on the current observation vector, with the aim to drive the robot toward the goal while safely and socially navigating around pedestrians and other obstacles.
\section{Simulation Environment}
\label{sec:env}
\begin{figure}[pos=!t]
    \centering
    \includegraphics[width=\linewidth,height=0.35\textheight,keepaspectratio]{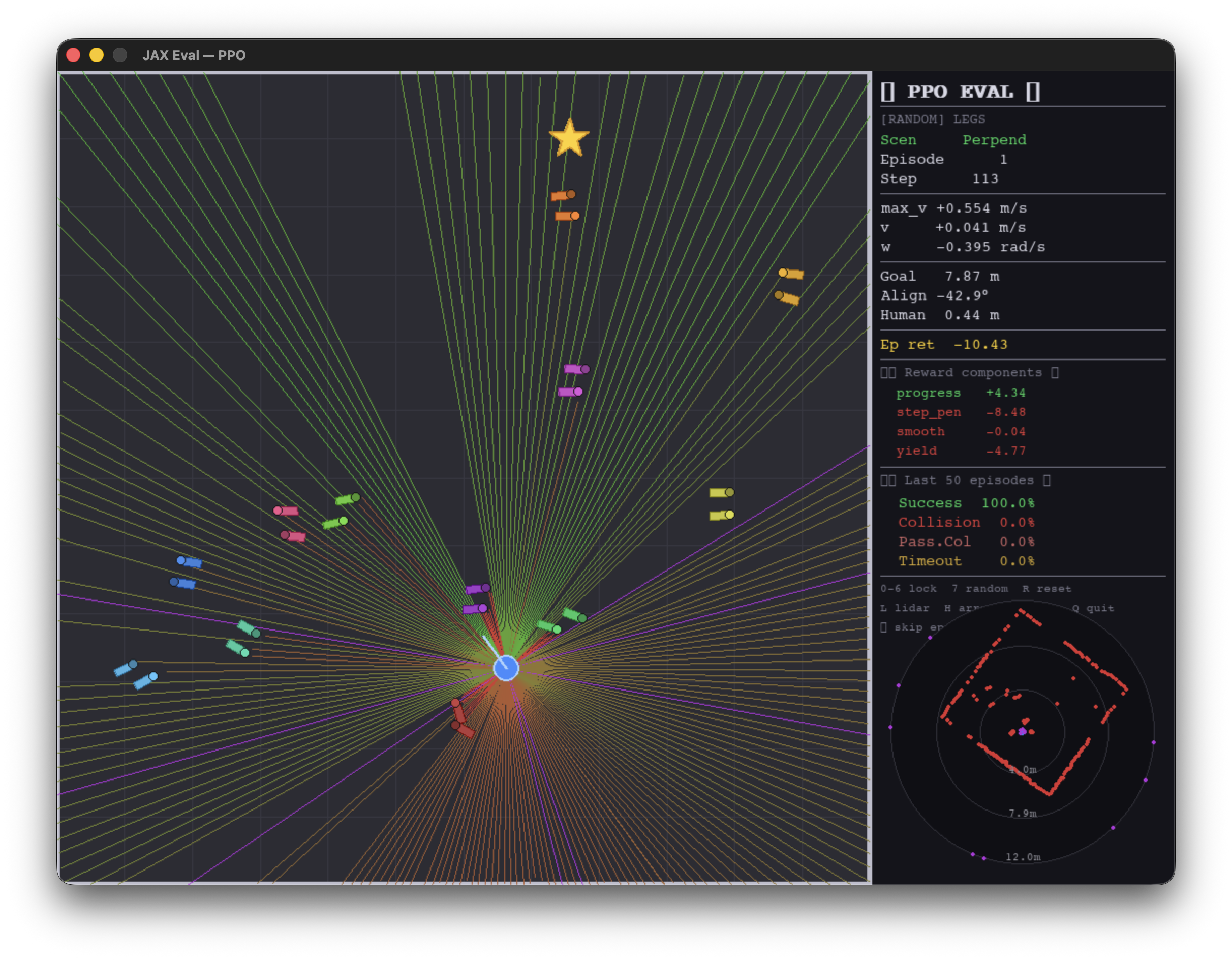}
    \caption{Visualization of the LegNav 2D simulation environment. The robotic agent (blue disc) navigates toward the goal (yellow star) using simulated 2D LiDAR (radial lines) to avoid dynamic human pedestrians, which are modeled as colored shoe pairs. The right panel tracks real-time quantities, evaluation metrics and reward components.}
    \label{fig:2denv}
\end{figure}

This section describes the LegNav 2D simulation environment (Fig.~\ref{fig:2denv}), which combines a noisy LiDAR sensor model, pedestrian dynamics governed by the HSFM, and the NSG model which determines the leg positions observed by the sensor.

\subsection{LiDAR Sensor Model}
\label{sec:lidar}

A 2D LiDAR sensor is employed featuring $L$ rays that span a full $360^\circ$ field of view up to a maximum range of $d_{\max}$. To simulate real-world conditions, Gaussian distance noise with a standard deviation of $\sigma$ is applied to all rays. Additionally, salt-and-pepper noise randomly affects $6\%$ of the rays, with half of these corrupted rays forced to the maximum range and the other half forced to zero.
In the LegNav simulator, the LiDAR sensor ray tracing model detects intersections with four obstacle types: room walls, circular obstacles, rectangular obstacles, and human legs, which are represented as a pair of small discs of radius $r_\ell$. This means that the policy observes the same dynamic scenario a physical ankle-height 2D LiDAR sensor would see: two small, moving clusters per pedestrian that alternately advance and pause as the person walks.

\subsection{Human Dynamics: Headed Social Force Model}
\label{sec:sfm}

Pedestrian dynamics are simulated using the HSFM \cite{farina2017walking}, which represents pedestrians as discs of radius $r_\text{body}$. Unlike the standard SFM \cite{helbing1995social}, the HSFM decouples body orientation from velocity, introducing torque-based dynamics to simulate the non-holonomic motion of real humans. At any given time step, the spatial state of a pedestrian is defined by their center of mass position $\mathbf{p}_{\text{body}}$, instantaneous velocity vector $\mathbf{v}_{\text{body}}$, and heading angle $\alpha_{\text{body}}$. Agents navigate toward waypoints driven by a goal-seeking force, while simultaneously evaluating repulsive forces from other pedestrians and static obstacles to avoid collisions. 
The net sum of these environmental forces generates both a longitudinal driving force and a rotational torque, which acts as an elastic spring-damper system that smoothly aligns the pedestrian's heading $\alpha_{\text{body}}$ with the direction of the net force. As a result, the HSFM provides a physically plausible, continuous trajectory and an instantaneous velocity $\mathbf{v}_{\text{body}}$ for each agent. This velocity influences the bipedal kinematics described below.

\subsection{Non-Slip Gait Model}
\label{sec:gait-model}
In the LegNav simulator, the pedestrian's feet are represented by a pair of rectangular shoes of length $L_{\text{shoe}}$, which are used to evaluate human-robot collisions. To mimic real-world sensor data, the pedestrian's legs are shaped as discs of radius $r_\ell$ centered at the shoe heel. These discs serve as the features detected by the LiDAR sensor.

Rather than treating pedestrians as particles, the explicit kinematics of two-legged walking is modeled. To do this, we introduce the Non-Slip Gait (NSG) model. Traditionally, a common approach for animating pedestrian legs in simulation expresses foot positions as sinusoidal offsets from the body center, oscillating at a cadence proportional to walking speed \cite{boulic1990global}. However, this formulation suffers from a kinematic inconsistency known as \emph{foot skating} (or foot-sliding): stance feet translate rigidly with the body rather than remaining planted, generating unrealistic gait dynamics. This artifact is heavily documented when superimposing bipedal mechanics onto standard center-of-mass crowd trajectories~\cite{zou2020reducing, beacco2015footstep}.

To avoid unrealistic behavior in LegNav, each foot is explicitly anchored to its last touchdown position during stance and it is free to swing only during the active half-cycle. Let $\mathbf{f}_L, \mathbf{f}_R \in \mathbb{R}^2$ represent the world-space current foot positions (the centers of the shoe rectangles), $\phi \in [0,1)$ the continuous gait phase variable and $\beta_L, \beta_R$ the left and right foot headings. All these quantities evolve over time; to keep the notation light,
we write the explicit dependence on $t$ only when an equation relates values at different time instants and we drop it otherwise\footnote{robot quantities use subscript notation $\cdot_t$~; pedestrian gait quantities use $\cdot(t)$~.}.

One complete gait cycle ($\phi \in [0, 1)$) consists of two symmetric half-periods; thus, if $\phi \in [0, 0.5)$, the left foot is planted and the right foot is swinging forward; if $\phi \in [0.5, 1.0)$, the right foot is planted and the left foot is swinging.
The gait phase $\phi$ advances at a rate $c$, the cadence, proportional to the body speed. The cadence and phase update are expressed as:
\begin{equation}
\label{eq:cadence}
\begin{aligned}
c &= f_{\max}~ \text{clip}\!\left(\frac{|\mathbf{v}_{\text{body}}|}{v_{\text{ref}}}, 0.3, 1.0\right) \\
\phi(t+1) &= \left(\phi(t) + c(t)~ \Delta t\right) \bmod 1.0
\end{aligned}
\end{equation}
The cadence is the number of left-right step pairs per second. It is proportional to the magnitude of the pedestrian's current velocity $|\mathbf{v}_{\text{body}}|$ from HSFM, normalized by a reference walking speed $v_{\text{ref}}$ and scaled by a maximum stepping frequency $f_{\max}$. The velocity ratio is bounded via a clip function: the lower bound prevents unrealistically slow stepping at near-zero speeds, while the upper bound caps the stepping frequency so that pedestrians moving faster than $v_{\text{ref}}$ do not increase cadence indefinitely. This reflects the empirical observation that below or above a natural cadence range, walking speed is regulated by decreasing or increasing the stride length rather than the stepping frequency~\cite{fukuchi2019effects}.

\begin{figure}[pos=!tp]
     \centering
     \begin{subfigure}[b]{\columnwidth}
         \centering
         \includegraphics[width=\linewidth,height=0.25\textheight,keepaspectratio]{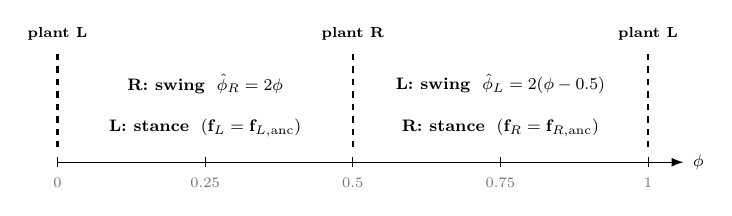}
         \caption{}
         \label{fig:gait_timeline}
     \end{subfigure}
     \vspace{0.5em}
     \begin{subfigure}[b]{\columnwidth}
         \centering
         \includegraphics[width=0.9\linewidth,height=0.25\textheight,keepaspectratio]{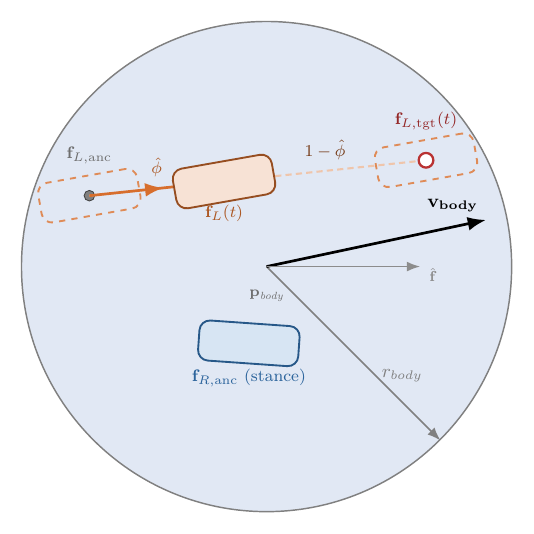}
         \caption{}
         \label{fig:gait_image}
     \end{subfigure}
    \caption{NSG model. (a) The gait phase $\phi \in [0,1)$ advances continuously; at each half-cycle crossing ($\phi = 0$ and $\phi = 0.5$) the corresponding foot plants and its anchor $\mathbf{f}_{\text{anc}}$ is updated. (b) During swing, the foot linearly interpolates from its anchor toward a target $\mathbf{f}_{\text{tgt}}$.}
\end{figure}

The gait cycle is driven by the gait phase variable $\phi \in [0,1)$.
Since only one foot swings during each half cycle, the gait phase is mapped to the progress of each foot swing $\hat{\phi} \in [0,1)$, which runs from $0$ at
lift-off toward $1$ at touchdown (see Fig.~\ref{fig:gait_timeline}), in the following way:
\begin{align}
\hat{\phi}_R &= 2\phi \quad \text{for } \phi \in [0,0.5),
\\
\hat{\phi}_L &= 2(\phi - 0.5) \quad \text{for } \phi \in [0.5,1.0).
\end{align}

\begin{figure}[pos=!t]
    \centering
    \includegraphics[width=\linewidth,height=0.3\textheight,keepaspectratio]{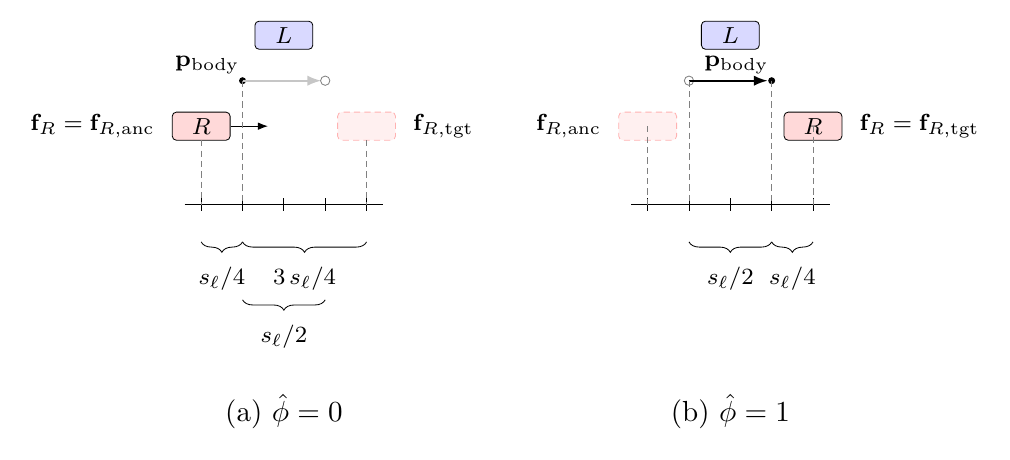}
    \caption{Swing-foot target offset at (a) the start ($\hat{\phi}=0$) and (b) the end 
    ($\hat{\phi}=1$) of a step. Distances are measured along the walking 
    direction from the current body center $\mathbf{p}_{\text{body}}$. In (a), at 
    lift-off, the foot starts $s_\ell/4$ behind the center and is targeted 
    $3s_\ell/4$ ahead. As the body advances by $s_\ell/2$ during the swing, 
    the same landing point ends up only $s_\ell/4$ ahead at touchdown, as shown in (b).}
    \label{fig:gait_coefficient}
\end{figure}

As shown in Fig.~\ref{fig:gait_image}, the swinging foot position
$\mathbf{f}(t)$ is obtained by using the swing progress
$\hat{\phi}$, to interpolate the planted anchor $\mathbf{f}_{\text{anc}}$ and a
dynamically updated target $\mathbf{f}_{\text{tgt}}(t)$:
\begin{equation}
\mathbf{f}(t)
=
\mathbf{f}_{\text{anc}}\,(1-\hat{\phi})
+
\mathbf{f}_{\text{tgt}}(t)\,\hat{\phi}.
\end{equation}
The anchor $\mathbf{f}_{\text{anc}}$ is the foot position at the end of the previous step. The target $\mathbf{f}_{\text{tgt}}(t)$
is the point the foot moves toward during the swing.
Such a target is dynamically updated as
\begin{align}
\mathbf{f}_{R,\text{tgt}}(t) 
&= \mathbf{p}_{\text{body}}(t) + \hat{\mathbf{f}}(t)\,\ell(\hat{\phi}_R) - \hat{\mathbf{l}}(t)\frac{w_{\text{hip}}}{2}, \nonumber \\
&\hspace{4em} \phi \in [0,0.5), \label{eq:right_foot} \\[0.5ex]
\mathbf{f}_{L,\text{tgt}}(t) 
&= \mathbf{p}_{\text{body}}(t) + \hat{\mathbf{f}}(t)\,\ell(\hat{\phi}_L) + \hat{\mathbf{l}}(t)\frac{w_{\text{hip}}}{2}, \nonumber \\
&\hspace{4em} \phi \in [0.5,1.0), \label{eq:left_foot}
\end{align}
where $\hat{\mathbf{f}}(t)$ is the unit vector
aligned with $\alpha_{\text{body}}(t)$ and $\hat{\mathbf{l}}(t)$ the unit vector
orthogonal to $\hat{\mathbf{f}}(t)$. The lateral offset
$\pm\hat{\mathbf{l}}(t)\,w_{\text{hip}}/2$ places the two feet symmetrically
about the body heading, separated by the hip width $w_{\text{hip}}$, while the
forward offset $\hat{\mathbf{f}}(t)\,\ell(\hat{\phi})$ sets how far ahead of the
body the foot goes. The
forward reach is determined by the step-length function
\begin{equation}
\ell(\hat{\phi}) =
\left(
\frac{3}{4} - \frac{1}{2}\hat{\phi}
\right)
s_\ell
,
\qquad
s_\ell = \frac{|\mathbf{v}_{\text{body}}|}{c},
\end{equation}
where $s_\ell$ approximates the \emph{stride length}, i.e., the distance the
body covers over one full left--right cycle.
The swing process is illustrated in Fig.~\ref{fig:gait_coefficient}. Consider walking in a straight line at constant speed. Over one full cycle both the body and each foot advance by a stride $s_\ell$. Within a single step, however, the body moves only $s_\ell/2$, while the swinging foot covers a distance $s_\ell$ and overtakes the body. Because the target is always measured from the \emph{current} body center, its position changes during the swing. At lift-off ($\hat{\phi}=0$) the swing foot starts $s_\ell/4$ \emph{behind} the center; after moving one stride forward it lands $-s_\ell/4 + s_\ell = 3s_\ell/4$ ahead of the center. By touchdown ($\hat{\phi}=1$) the body advances by $s_\ell/2$. Hence the landing point is now only $(3/4 - 1/2)s_\ell = s_\ell/4$ ahead of the body center.

It is worth remarking that the targets are updated at every simulation timestep because the body can turn or change speed while a foot is still in
the air. A target fixed at lift-off would leave the foot landing in a position
no longer consistent with the body's pose, producing a visible
misalignment; updating it dynamically lets the swing trajectory follow the body's turns and speed variations. 

Accurately modeling the physical footprints of pedestrians also requires tracking the yaw orientation of each foot.
The left and right foot orientation $\beta_L$ and $\beta_R$ follow a stance-dependent update rule. During the \emph{stance} phase, the foot is planted on the ground and its orientation $\beta$ is frozen at its touchdown value. During the \emph{swing} phase, the foot is lifted from the ground and its orientation continuously tracks the body heading $\alpha_{\text{body}}$ provided by the HSFM. Formally, for the left foot:
\begin{equation}
\beta_L(t) = 
\begin{cases} 
\beta_L(t-1) & \text{if } \phi \in [0, 0.5) \text{ (stance)} \\
\alpha_{\text{body}}(t) & \text{if } \phi \in [0.5, 1.0) \text{ (swing)}
\end{cases}
\end{equation}
The same logic applies symmetrically to the right foot. This mechanism ensures that a foot rotates in the air during turns and lands aligned with the pedestrian's updated heading.

\section{Navigation Policy}
\label{sec:navigation_policy}

This section presents the navigation policy: the end-to-end CALF architecture $\pi(\mathbf{a}_t|\mathbf{o}_t)$, the DRL algorithms used to train it and the reward shaping for the RL framework.

\begin{figure*}[pos=!t]
\centering
\includegraphics[width=\textwidth,height=0.35\textheight,keepaspectratio]{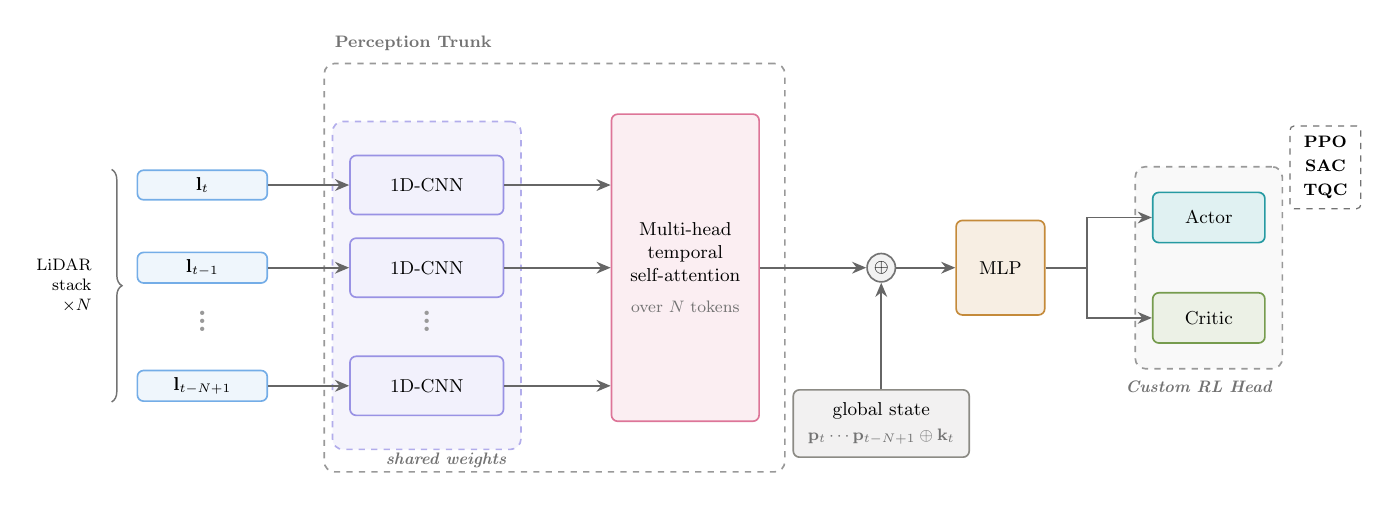}
\caption{CALF end-to-end policy architecture processing stacked 2D LiDAR frames via a shared CNN and temporal self-attention, fused with the global state. The shared representation is routed to a custom RL head containing actor and critic branches, which are structurally adapted based on the chosen RL algorithm.}
\label{fig:nn}
\end{figure*}

\subsection{CALF Architecture}
\label{subsec:calf_architecture}
The observation vector encodes heterogeneous information (see (\ref{eq:observations})) which requires specialized processing streams.
For this reason, we designed CALF (``\textit{Convolutional Attention for Leg Features}''), a novel end-to-end hybrid network architecture that combines 1D-CNNs, multi-head temporal self-attention and MLP modules (Fig.~\ref{fig:nn}).
The LiDAR stream stack (\ref{eq:observations}) is handled by a \textit{Perception Trunk} composed of a convolutional encoder and a temporal self-attention module.
A 1D-CNN Encoder with three layers processes each of the $N$ LiDAR scans independently, applying the same learned weights to every frame in the temporal stack and ensuring consistent feature extraction. 
All three layers use a stride greater than one, so the scan is progressively shortened and each successive layer aggregates a wider sector of the original LiDAR scan. As a result, deeper features summarize regions large enough to capture the geometric patterns, in particular the range returns produced by human legs.

After the three convolutional layers, the spatial dimension is flattened and projected to a 64-dimensional feature vector, producing one compact token per frame.
The $N$ feature vectors coming from the 1D-CNN are processed by a single-layer multi-head temporal self-attention module with four heads, which captures environmental dynamics across the frame stack. 
Employing multiple heads allows the network to concurrently attend to diverse environmental patterns within distinct representation subspaces. For instance, one head may specialize in tracking fast-moving pedestrians, while another monitors static boundaries or ego-motion. By computing attention across the temporal stack, the network autonomously correlates past and present spatial features and implicitly extracts dynamic states.
The \textit{global state} vector (Fig.~\ref{fig:nn}), containing the concatenated goal stack and kinematic vector, bypasses the perception trunk and is fed directly into the fusion stage without any dedicated feature extraction.
There, it is concatenated with the spatio-temporal LiDAR features produced by the attention module, forming a unified multi-modal representation. This combined input is processed by a two-layer MLP that produces the final shared embedding used by both the actor and critic heads.
The shared embedding is routed to the actor and critic branches (\textit{Custom RL Head} in Fig.~\ref{fig:nn}), whose internal structure depends on the RL algorithm used for training, as detailed in Section~\ref{sec:algorithms}.

The CALF network parameters are reported in Table~\ref{tab:network-params}.

\begin{table}[pos=!t]
\caption{CALF network parameters.}
\label{tab:network-params}
\centering
\footnotesize
\setlength{\tabcolsep}{3pt}
\begin{tabular}{p{0.24\columnwidth}p{0.40\columnwidth}r}
\toprule
\textbf{Block} & \textbf{Parameter} & \textbf{Value} \\
\midrule
\multirow{4}{0.22\columnwidth}{\raggedright 1D-CNN Encoder}
 & Filters per layer  & 32, 64, 64 \\
 & Kernel sizes       & 7, 5, 3 \\
 & Stride (all layers) & 2 \\
 & Output token dim.  & 64 \\
\midrule
\multirow{3}{0.22\columnwidth}{\raggedright Multi-head temporal Self-attention}
 & Attention layers        & 1 \\
 & Heads         & 4 \\
 & Dim. per head & 16 \\
\midrule
MLP & Hidden widths & 256, 128 \\
\midrule
\multirow{2}{0.22\columnwidth}{\raggedright Actor / Critic}
 & Actor output dim. & 2 \\
 & Critic hidden width & 64 \\
\bottomrule
\end{tabular}
\end{table}

\subsection{RL Methods}
\label{sec:algorithms}
The heads of the CALF network depend on the DRL method employed for training. 
In this work we adopt three different algorithms: PPO \cite{schulman2017proximal}, SAC \cite{haarnoja2018soft} and TQC \cite{kuznetsov2020controlling}.
All three algorithms fall under the class of \textit{model-free Actor-Critic} methods \cite{konda1999actor}, in which two separate components are trained simultaneously: the Actor network represents the policy, mapping observations to actions.
The Critic network estimates the value function, which guides the Actor's improvement.

PPO has been chosen due to its robustness and strong empirical performance across a wide range of RL tasks \cite{schulman2017proximal, henderson2018deep}. It belongs to the family of policy gradient methods and optimizes a clipped surrogate objective to limit policy updates. This mechanism ensures stable learning by preventing drastic changes in policy between updates. While PPO is known for its simplicity and reliability, it often underperforms in environments requiring deeper exploration or featuring high stochasticity, as it lacks an explicit mechanism for entropy-based exploration \cite{rajeswaran2017towards, isele2018navigating}.

SAC, in contrast, is an off-policy algorithm that introduces entropy regularization into the objective function \cite{haarnoja2018soft, haarnoja2018applications}. This encourages the policy to explore more diverse actions by maximizing a combination of expected reward and entropy. As a result, SAC tends to learn smoother, more adaptable behaviors and performs particularly well in environments with high variability or uncertainty. Its stochastic nature allows it to avoid converging prematurely to suboptimal deterministic policies, which can be a limitation in methods like PPO \cite{haarnoja2019learning, ma2022sacnavigation}.

TQC builds upon SAC by incorporating advances from distributional RL \cite{dabney2018distributional, kuznetsov2020controlling}. 
Specifically, it models the action–value distribution using quantile regression and truncates the largest quantiles of the target, inducing a controlled pessimism that reduces overestimation bias, a common problem in value-based RL methods. TQC can be particularly advantageous in safety-critical navigation tasks, where overly aggressive actions can result in failures or unsafe behaviors \cite{bellemare2017distributional, chen2021robotnav}.  Its application in mobile robot navigation has been explored only recently \cite{bae2026deep, choton2025efficient}.

Depending on the chosen RL algorithm, the shared network trunk is connected to different algorithm-specific heads.
For all three algorithms, at training time the actor outputs a mean 
$\boldsymbol{\mu}(\mathbf{o}_t) \in \mathbb{R}^2$ and a standard deviation $\boldsymbol{\sigma} \in \mathbb{R}^2$, from 
which a sample $\mathbf{w}_t = [w_v, w_\omega]^\top \in 
\mathbb{R}^2$ is drawn:
\begin{equation}
\mathbf{w}_t \sim \mathcal{N}\!\left(\boldsymbol{\mu}(\mathbf{o}_t),\, 
\operatorname{diag}\!\left(\boldsymbol{\sigma}^2\right)\right),
\end{equation}
and the physical action $\mathbf{a}_t$ is obtained by squashing the sample $\mathbf{w}_t$ via $\tanh$, to comply with the action space bounds (\ref{eq:action_space}). 

The two algorithm families differ in how $\boldsymbol{\sigma}$ is 
parameterized: for \textit{PPO}, $\boldsymbol{\sigma}$ is a learned 
observation-independent parameter \cite{schulman2017proximal}, whereas for 
\textit{SAC} and \textit{TQC} the standard deviation 
$\boldsymbol{\sigma}(\mathbf{o}_t)$ is output directly by the actor, making 
the exploration level observation-dependent. For \textit{PPO}, the critic is an independent MLP head. 
\textit{SAC} and \textit{TQC} instead optimize an off-policy 
maximum-entropy objective and must therefore evaluate the exact policy density \cite{haarnoja2018soft}. 
For \textit{SAC}, the critic consists of two Q-networks whose minimum is used as the target \cite{haarnoja2018soft}. \textit{TQC} extends this design with distributional critics: an ensemble of networks, each outputting
quantile atoms, is adopted; the largest quantiles are truncated from the pooled target \cite{kuznetsov2020controlling}.

\subsection{Reward shaping}
\label{subsec:reward_shaping}
The reward function $r_t$ is the sum of several terms combining navigation objectives with social compliance incentives. The main per-step terms are:
\begin{align}
r_{\text{prog},~t}   &= \alpha_{\text{prog}} \cdot (d_{t-1} - d_t) \label{eq:r_prog} \\ 
r_{\text{step},~t}       &= -\alpha_{\text{step}} \label{eq:r_step} \\ 
r_{\text{smooth},~t}    &= -\alpha_{\text{smooth}}\,|\omega_t - \omega_{t-1}| \label{eq:r_smooth} \\ 
r_{\text{speed},~t}      &= \alpha_{\text{step}}\,\frac{v_t}{v_{\max}} \label{eq:r_speed} \\ 
r_{\text{heading},~t}    &= \alpha_{\text{heading}}\,\max(0,\cos\rho_t)\,\left(\frac{v_t}{v_{\max}}\right) \label{eq:r_heading} \\ 
r_{\text{comfort},~t}    &= -\alpha_{\text{comfort}}\sum_i\max\!\left(0,\, 1 -\frac{d_{i,t}}{d_{\text{comfort}}}\right) \nonumber \\
                         &\quad \cdot \left(1 + \frac{v_t}{v_{\max}}\right) \label{eq:r_comfort_penalty}
\end{align}
where $d_t$ is the Euclidean distance to the goal, $\rho_t$ is the angular error between the robot heading and the goal direction in the ego-frame, $d_{i,t}$ is the distance to the $i$-th pedestrian center (body frame), and $d_{\text{comfort}}$ is the personal space radius. Each term shapes a distinct aspect of the desired navigation behavior. The progress term $r_{\text{prog}}$ (\ref{eq:r_prog}) rewards reduction of the goal distance and provides the dense learning signal. The step penalty $r_{\text{step}}$ (\ref{eq:r_step}) imposes a constant cost per time step, discouraging idle behavior. The smoothness term $r_{\text{smooth}}$ (\ref{eq:r_smooth}) penalizes changes in angular velocity between consecutive steps, suppressing oscillatory steering. The speed term $r_{\text{speed}}$ (\ref{eq:r_speed}) mitigates the step penalty in proportion to the robot's forward velocity, incentivizing the agent to move at higher speeds. The heading term $r_{\text{heading}}$ (\ref{eq:r_heading}) rewards forward motion only when it is directed toward the goal, coupling the speed incentive to goal alignment. The comfort term $r_{\text{comfort}}$ (\ref{eq:r_comfort_penalty}) penalizes intrusion into pedestrian personal space, with the penalty growing as the distance $d_{i,t}$ falls below $d_{\text{comfort}}$; the factor $(1 + v_t/v_{\max})$ scales the penalty with speed, so that high-speed intrusions are penalized more than proximity at slow speed.
Additionally, a novel per-step reward term is proposed to encourage the
robot to yield when a pedestrian crosses its path. To this end, a frontal
\emph{yield zone} is defined as the circular sector of radius
$d_{\mathrm{yield}}$ and field of view angle $\psi_{\mathrm{FOV}}$, centered on the
robot and aligned with its heading. At each time step $t$, the pedestrians
occupying this sector are collected in the index set
\begin{equation}
\mathcal{Y}_t
=
\left\{
i \in \mathcal{P} :
d_{i,t} < d_{\mathrm{yield}},
~
|\psi_{i,t}| < \frac{\psi_{\mathrm{FOV}}}{2}
\right\},
\label{eq:yield_set}
\end{equation}
where $\mathcal{P}$ is the set of pedestrian indices, and $d_{i,t}$ and
$\psi_{i,t}$ denote the distance and bearing angle to the $i$-th pedestrian
relative to the robot heading. Whenever $\mathcal{Y}_t \neq \varnothing$,
an urgency scalar
\begin{equation}
u_t
=
\frac{
d_{\mathrm{yield}}
-
\min\limits_{i \in \mathcal{Y}_t} d_{i,t}
}
{d_{\mathrm{yield}}}
\in (0,1]
\end{equation}
quantifies how deeply the closest pedestrian has penetrated the zone,
growing from $0$ at the zone boundary towards $1$ as the pedestrian
approaches the robot centre. The yield term then rewards velocities below a threshold $v_\text{stop}$ stopping and
penalizes velocities above $v_\text{stop}$ in proportion to this urgency:
\begin{equation}
r_{\mathrm{yield},t}
=
\begin{cases}
\alpha_{\mathrm{yield}}\,u_t,
& \text{if } \mathcal{Y}_t \neq \varnothing
\text{ and } v_t \le v_{\mathrm{stop}},
\\[2mm]
-\alpha_{\mathrm{yield}}\,u_t\,v_t,
& \text{if } \mathcal{Y}_t \neq \varnothing
\text{ and } v_t > v_{\mathrm{stop}},
\\[2mm]
0,
& \text{otherwise}.
\end{cases}
\label{eq:reward_yield}
\end{equation}
The speed reward is suppressed while the zone is occupied,
$r_{\mathrm{speed},t} = 0$ if $\mathcal{Y}_t \neq \varnothing$, so that it
cannot counteract the stopping incentive.

The reward coefficients $\alpha$ are all positive and they can be tuned to trade off navigation efficiency against social compliance. Their values, selected empirically through preliminary training runs to balance the magnitude of individual reward terms, are listed in Table~\ref{tab:reward-parameters-complete}.

Terminal episode events define terminal reward terms, which usually take larger values (see Table~\ref{tab:reward-parameters-complete}). Reaching the goal yields the highest positive reward $B_{\text{goal}}$, while terminal failures are penalized in decreasing order of severity: active collision with a pedestrian ($B_{\text{act}}$), collision with static obstacles ($B_{\text{obs}}$), timeout event ($B_{\text{timeout}}$), and passive collision with a pedestrian ($B_{\text{pas}}$).

Pedestrian collisions are detected when the robot and a shoe-box come into contact, and are classified as \emph{active} or \emph{passive} according to whether the robot had the opportunity to yield. A contact with pedestrian $i$ at time $t$ is labelled active if the robot is moving above the stopping threshold ($v_t > v_{\mathrm{stop}}$) and that same pedestrian occupies the frontal yield zone, i.e. $i \in \mathcal{Y}_t$. Taken together, the two conditions indicate that the pedestrian was visible ahead of the robot, that the yield reward was actively demanding a stop, and that the robot nevertheless kept advancing into the pedestrian's path; such events are therefore penalised most severely.

Instead, any other robot--shoe contact is labelled passive. This covers the case in which the robot is nearly stationary ($v_t \le v_{\mathrm{stop}}$) and a pedestrian walks into it and the case in which the contacted pedestrian lies outside the frontal yield zone ($i \notin \mathcal{Y}_t$), such as a lateral or rear approach the robot could not reasonably anticipate. Since responsibility does not lie entirely with the robot in these situations, they receive a smaller penalty.

\section{Training}
\label{sec:training}

\begin{figure*}[pos=!t]
    \centering
    \begin{subfigure}[t]{0.21\linewidth}
        \includegraphics[width=\linewidth]{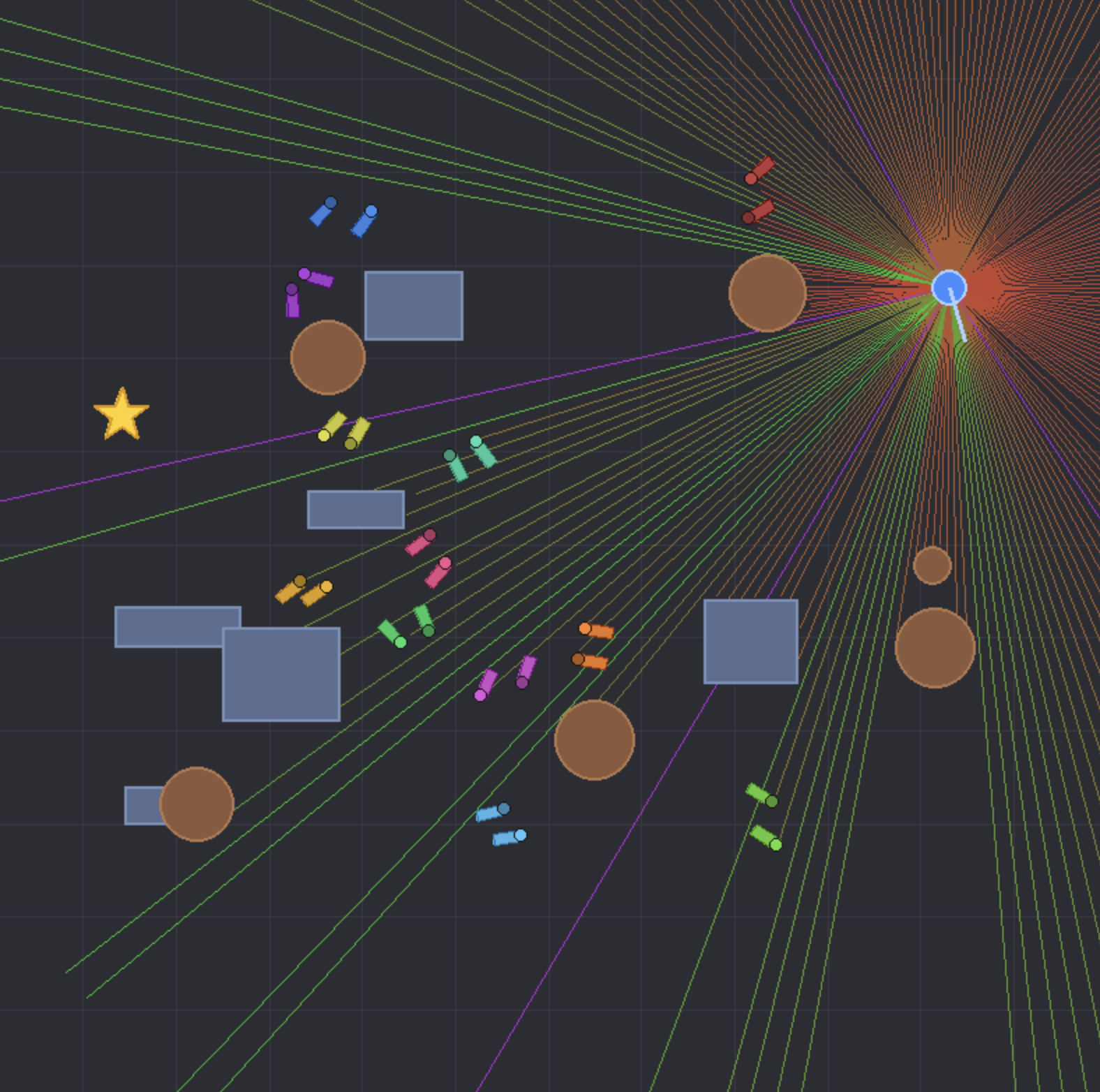}
        \caption{Random Scenario}
    \end{subfigure}\hfill
    \begin{subfigure}[t]{0.21\linewidth}
        \includegraphics[width=\linewidth]{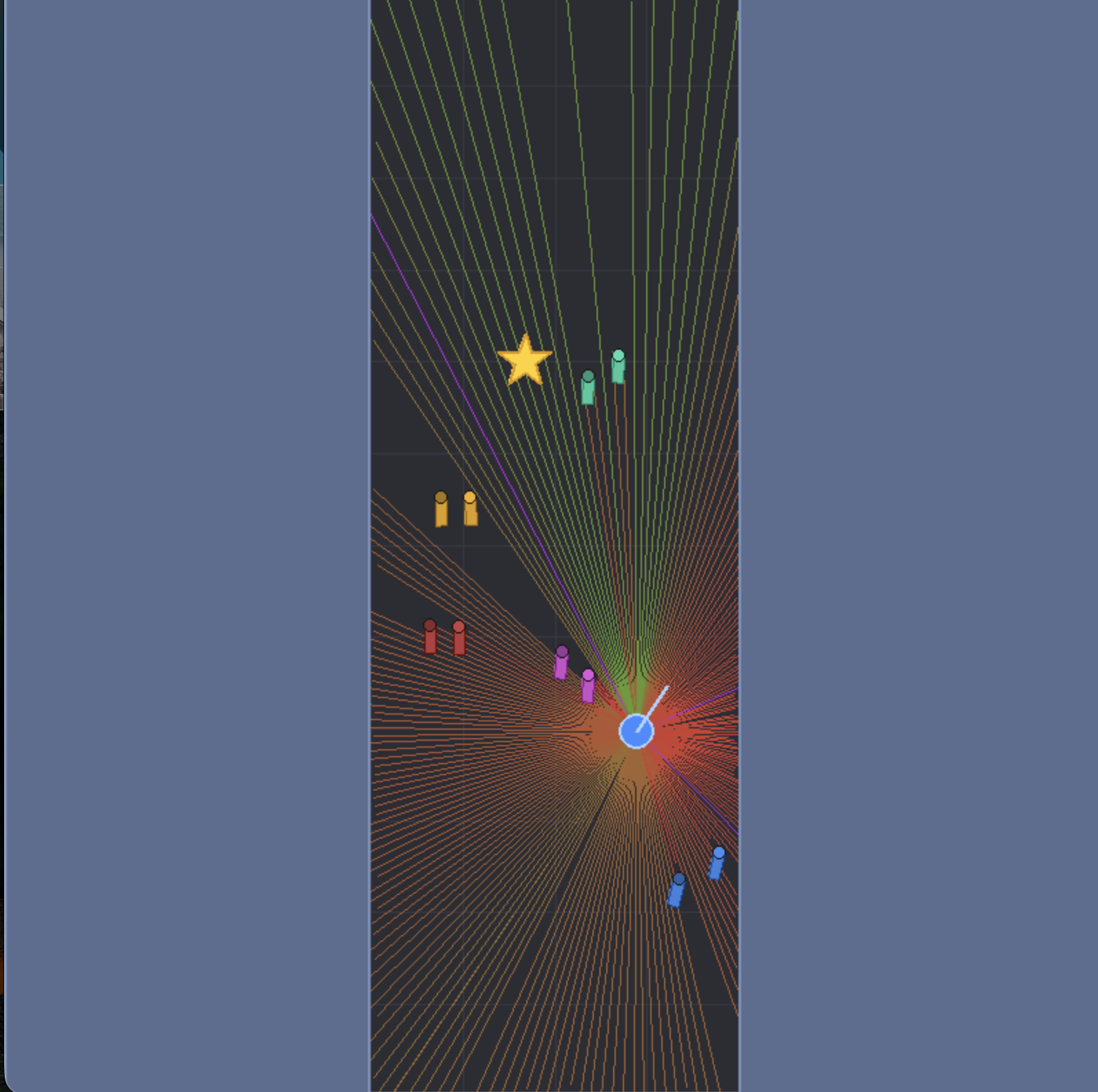}
        \caption{Parallel Traffic}
    \end{subfigure}\hfill
    \begin{subfigure}[t]{0.21\linewidth}
        \includegraphics[width=\linewidth]{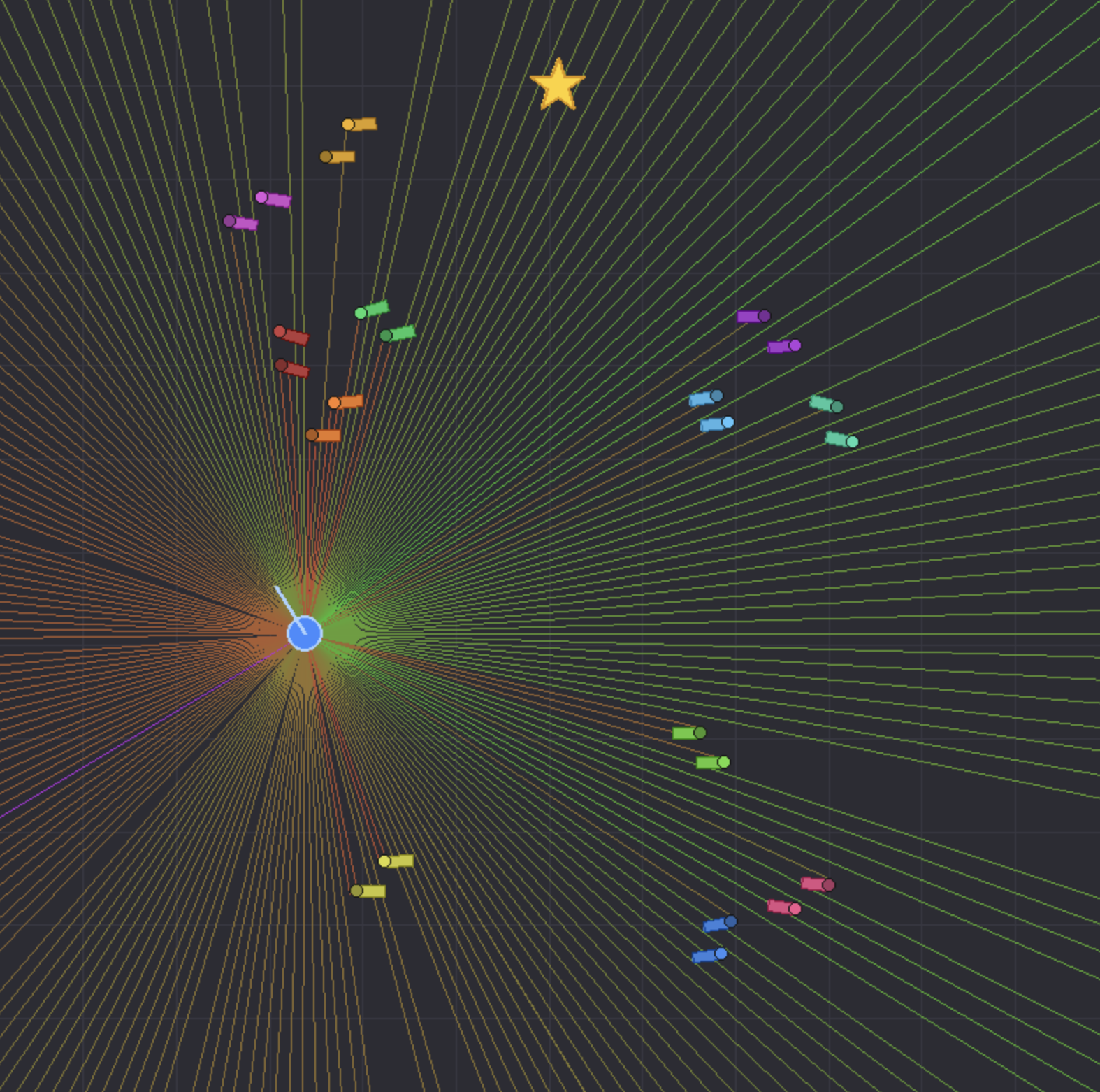}
        \caption{Perpendicular Traffic}
    \end{subfigure}\hfill
    \begin{subfigure}[t]{0.21\linewidth}
        \includegraphics[width=\linewidth]{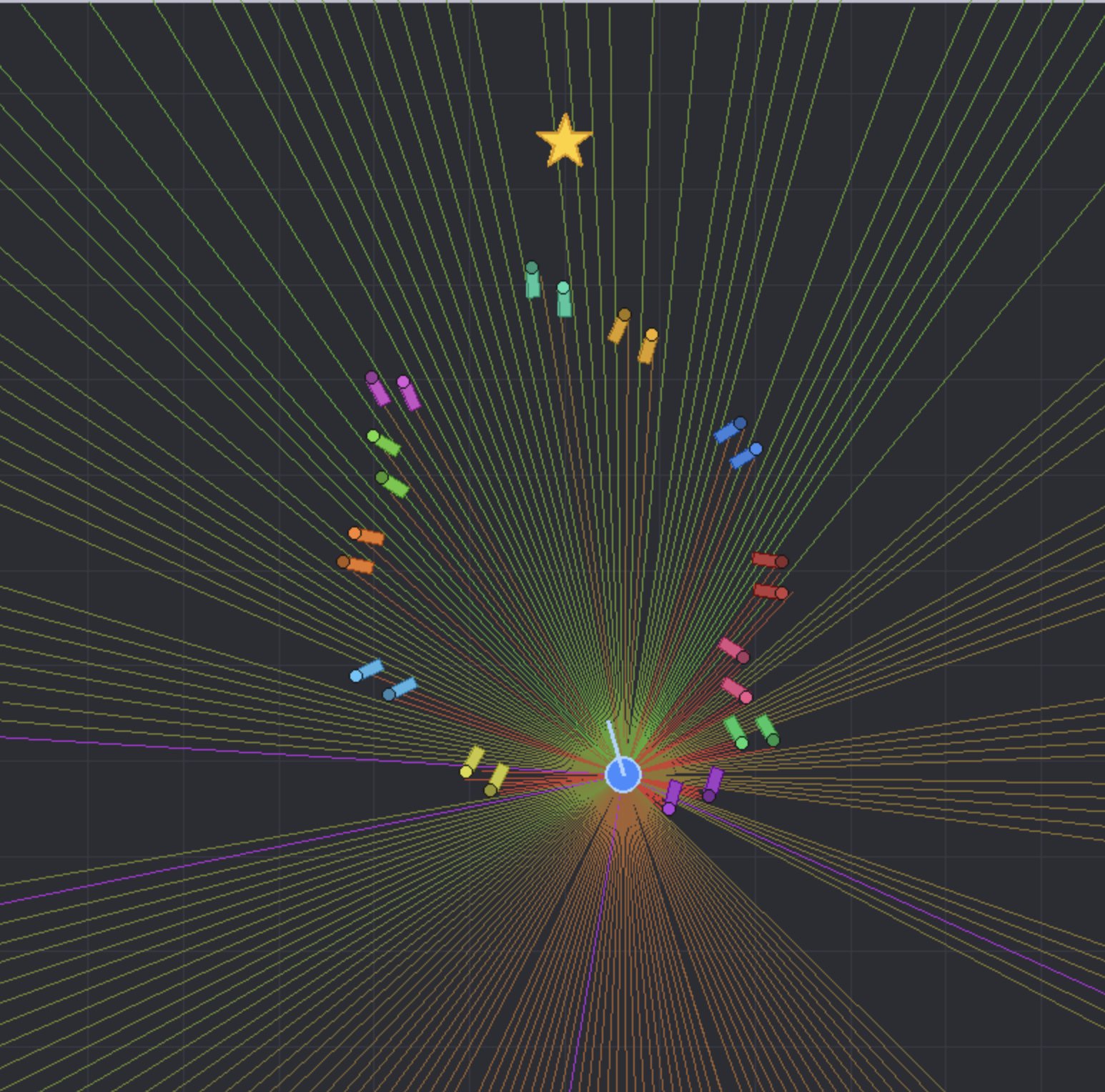}
        \caption{Circular Traffic}
    \end{subfigure}

    \vspace{0.8em}

    \begin{subfigure}[t]{0.21\linewidth}
        \includegraphics[width=\linewidth]{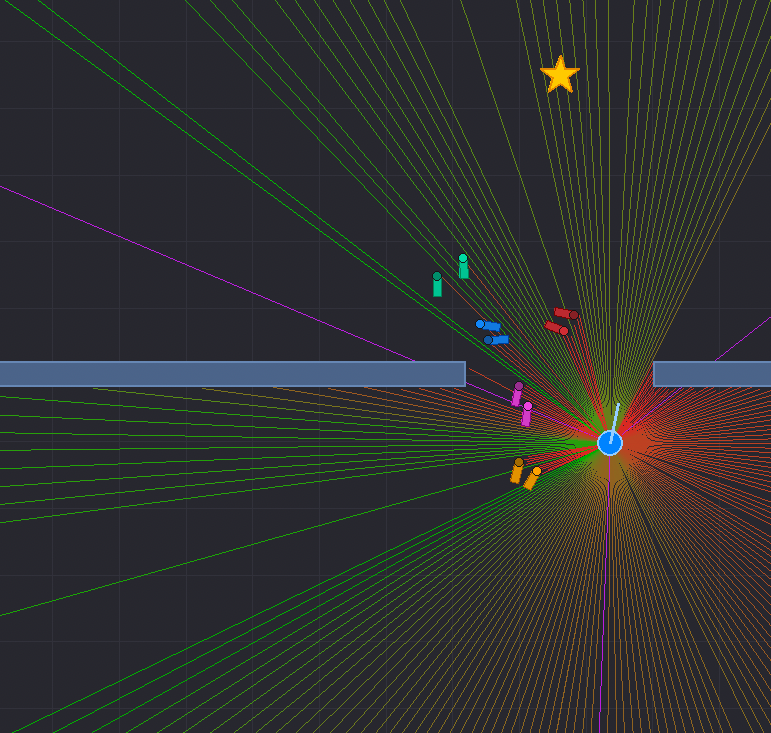}
        \caption{Bottleneck}
    \end{subfigure}\hfill
    \begin{subfigure}[t]{0.21\linewidth}
        \includegraphics[width=\linewidth]{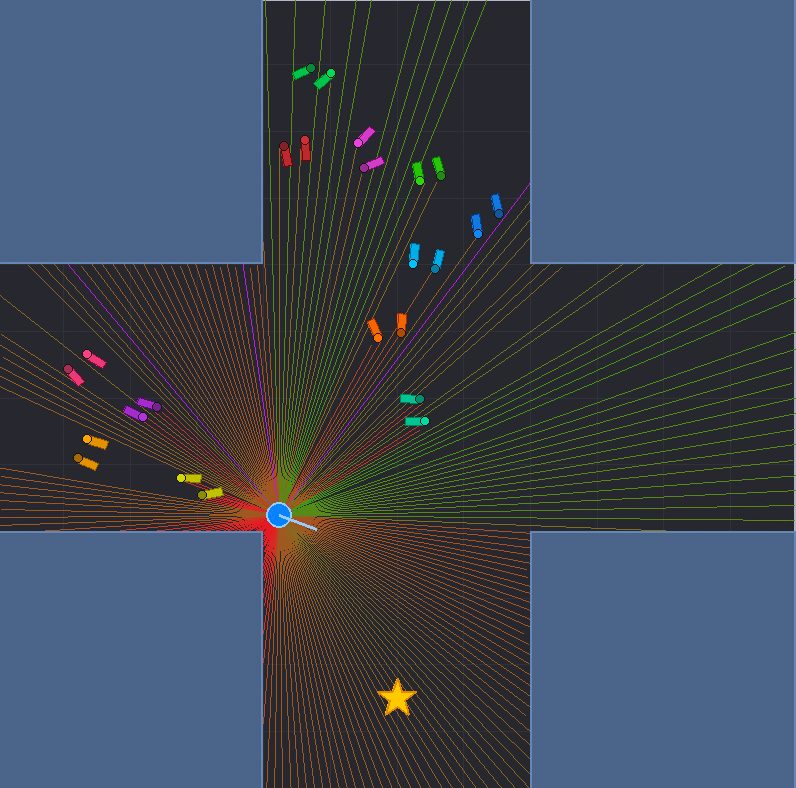}
        \caption{Intersection}
    \end{subfigure}\hfill
    \begin{subfigure}[t]{0.21\linewidth}
        \includegraphics[width=\linewidth]{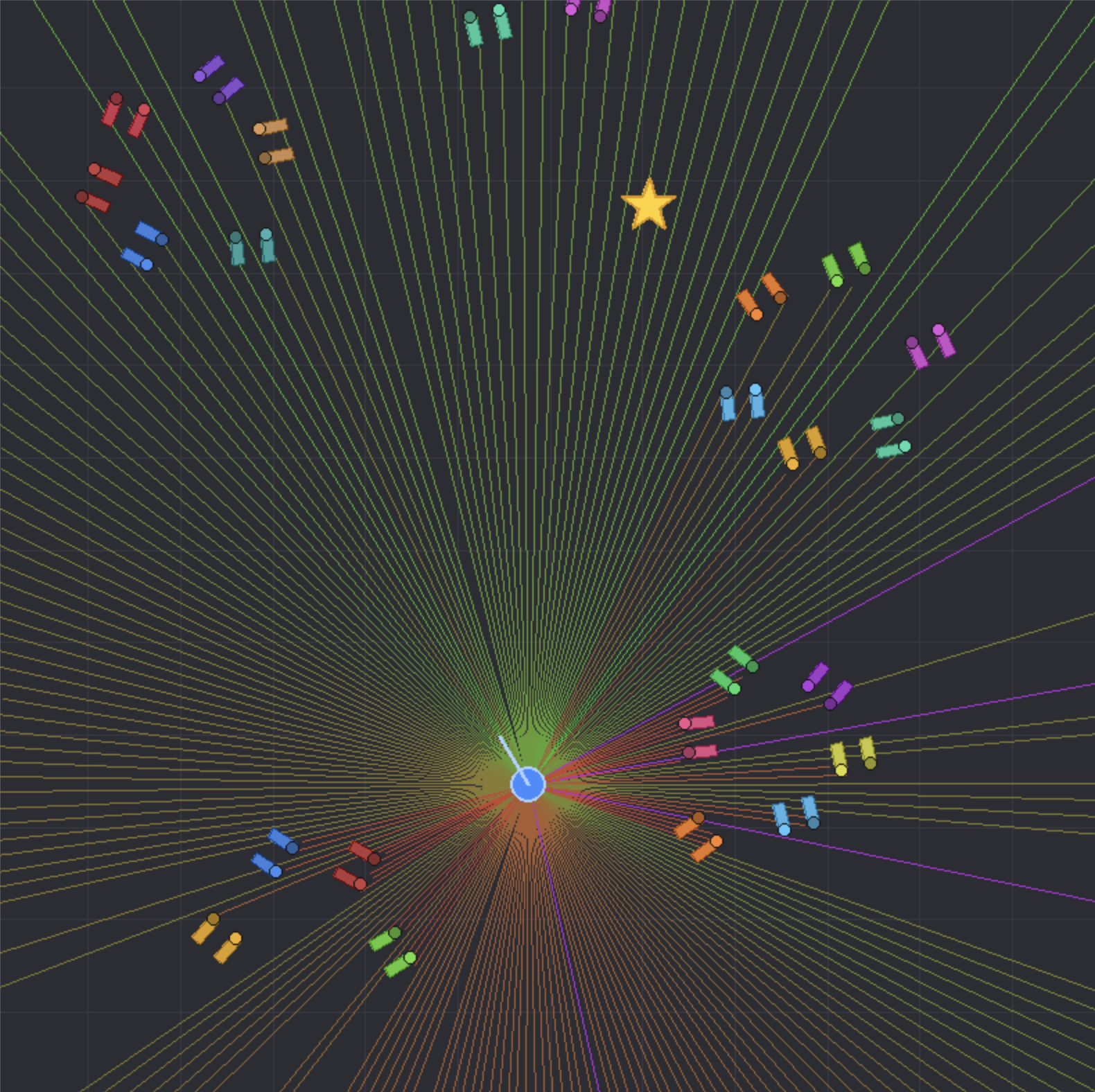}
        \caption{Static Groups}
    \end{subfigure}\hfill

    \caption{Scenarios used during training in simulation.}
    \label{fig:training_scenarios}
\end{figure*}

To ease policy generalization, the LegNav simulator contains seven training navigation scenarios (Fig.~\ref{fig:training_scenarios}) with varying crowd and static obstacle designs, capturing key contextual complexity factors, such as crowd directionality and environment geometry, as characterized by~\cite{stratton2025characterizing}. 

\begin{enumerate}
\item \textbf{Random Scenario}: randomized pedestrian positions, random static obstacles (discs and boxes), and random robot/goal placements. Serves as the general-purpose training scenario.
\item \textbf{Parallel Traffic (Corridor)}: a corridor in which pedestrians walk counter-directionally to the robot, creating oncoming traffic with constrained passing space.
\item \textbf{Perpendicular Traffic}: pedestrians traverse the room laterally while the robot navigates longitudinally, creating crossing conflict situations.
\item \textbf{Circular Traffic}: pedestrians are arranged on a circle and walk to diametrically opposite positions, creating a rotary intersection pattern.
\item \textbf{Bottleneck}: pedestrians approach a narrow gap in a horizontal wall, creating queuing and merging dynamics which require precise maneuvering.
\item \textbf{Intersection}: a cross-shaped open space in which pedestrians move along corridors in every direction while the robot must navigate through the intersection.
\item \textbf{Static Groups}: pedestrians stand still in groups placed in random positions.
\end{enumerate}

Within each scenario, we employ a domain randomization strategy similar to~\cite{ahsen2025domain}, where the start and goal positions of the robot, as well as the initial coordinates of the pedestrians, are randomized across training episodes to ensure robust policy learning.
Scenarios are progressively unlocked during training as a function of the curriculum success rate, as described later in this section.
The maximum linear speed $v_{\max}$ is sampled uniformly from $[0.2, 2.0]$~m/s, since we want the learned policies to be deployable on real mobile robots with different maximum linear speeds.

\begin{figure*}[pos=!t]
    \centering
    \begin{subfigure}[t]{0.27\linewidth}
        \includegraphics[width=\linewidth]{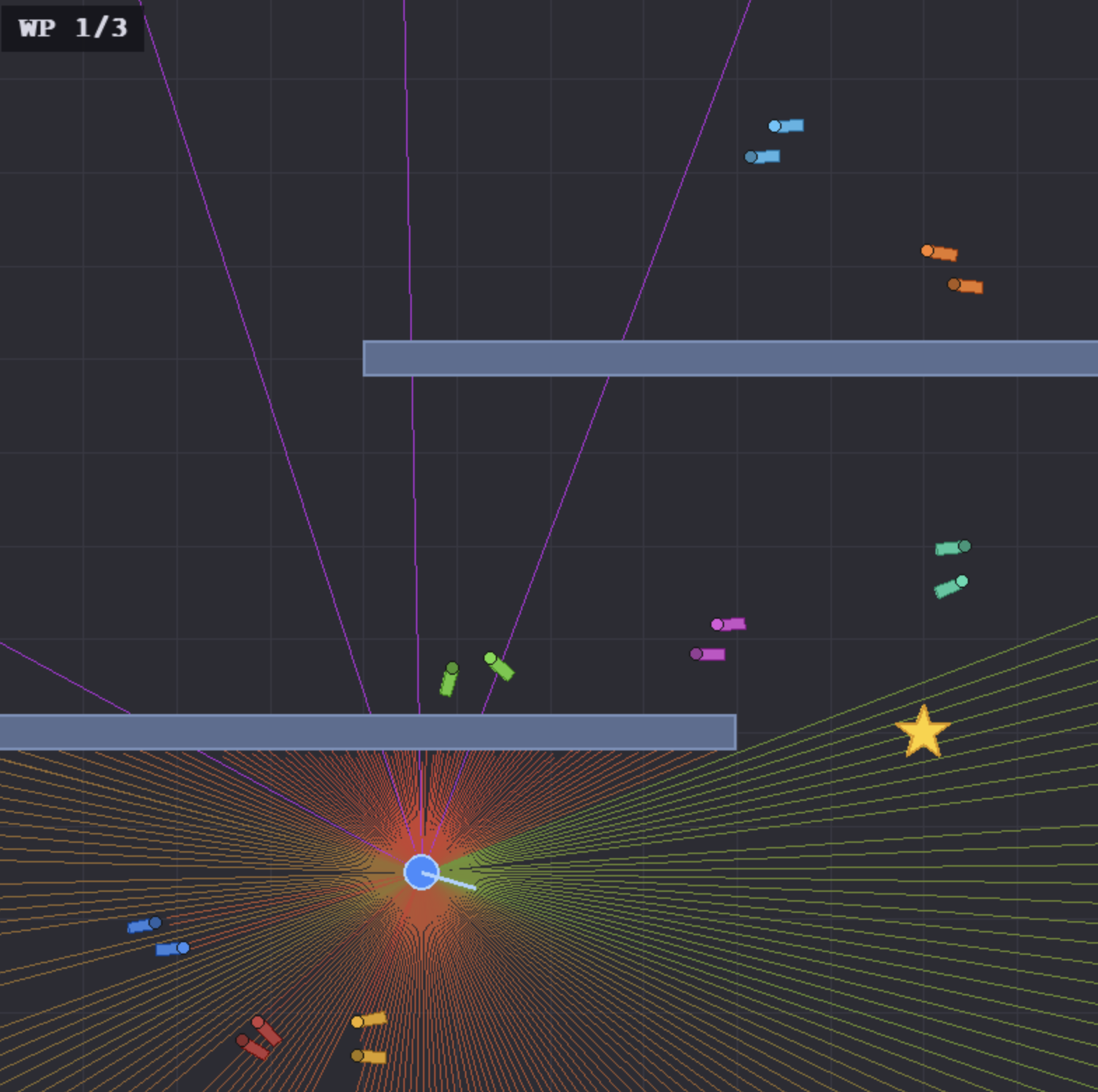}
        \caption{S-Corridor}
    \end{subfigure}\hfill
    \begin{subfigure}[t]{0.27\linewidth}
        \includegraphics[width=\linewidth]{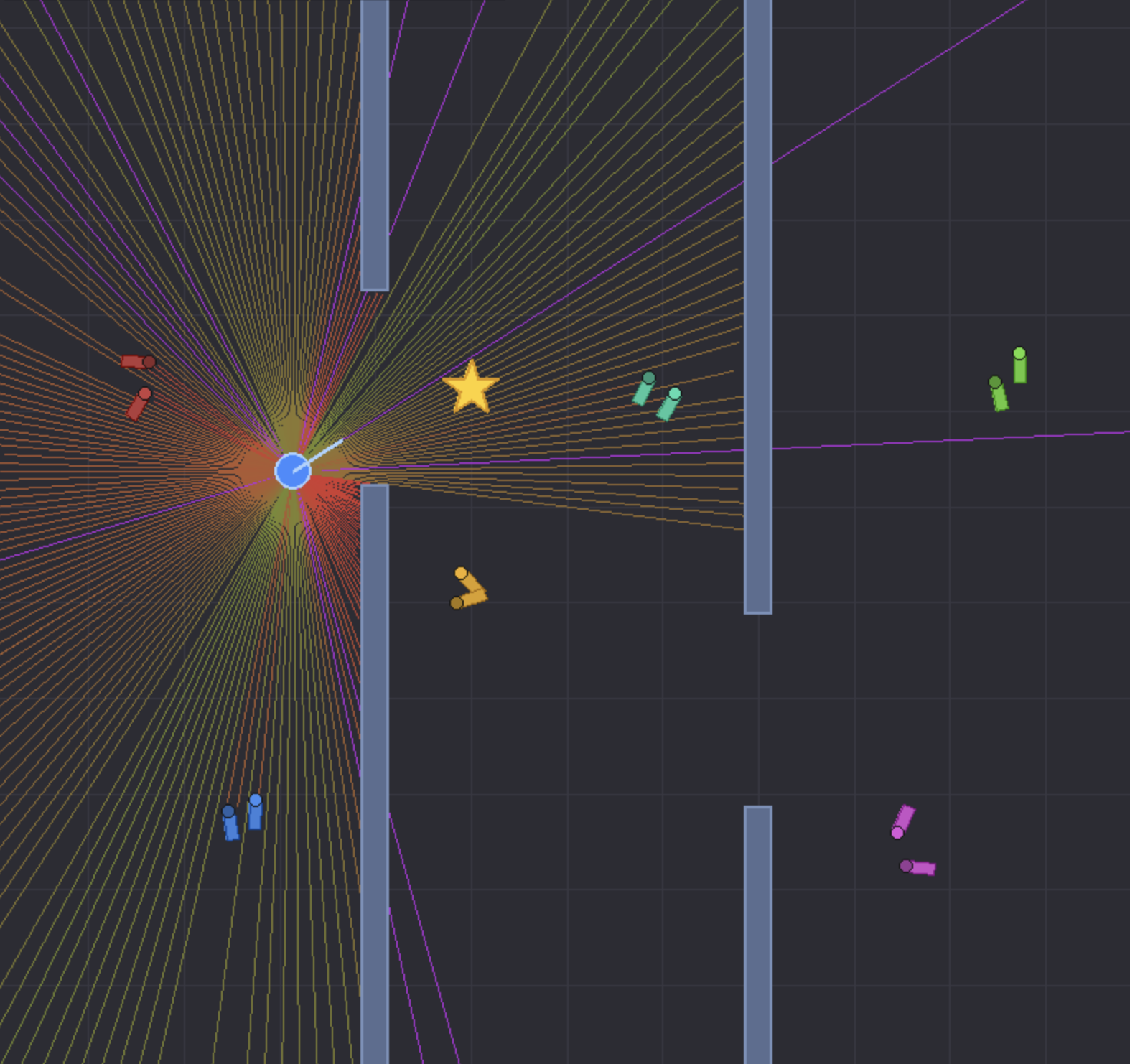}
        \caption{Sequential Rooms}
    \end{subfigure}\hfill
    \begin{subfigure}[t]{0.27\linewidth}
        \includegraphics[width=\linewidth]{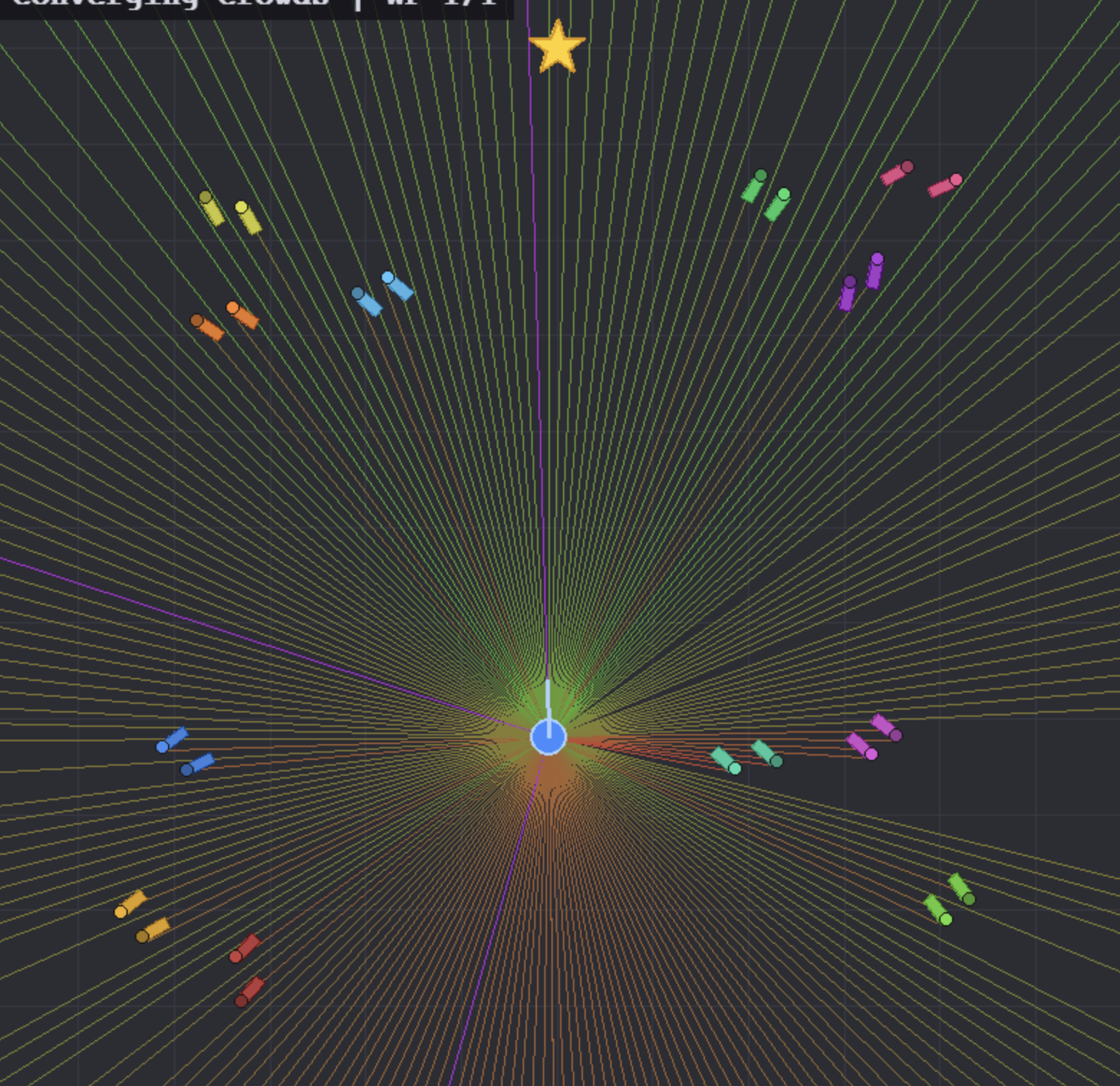}
        \caption{Converging Crowds}
    \end{subfigure}\hfill

    \vspace{0.8em}

    \begin{subfigure}[t]{0.27\linewidth}
        \includegraphics[width=\linewidth]{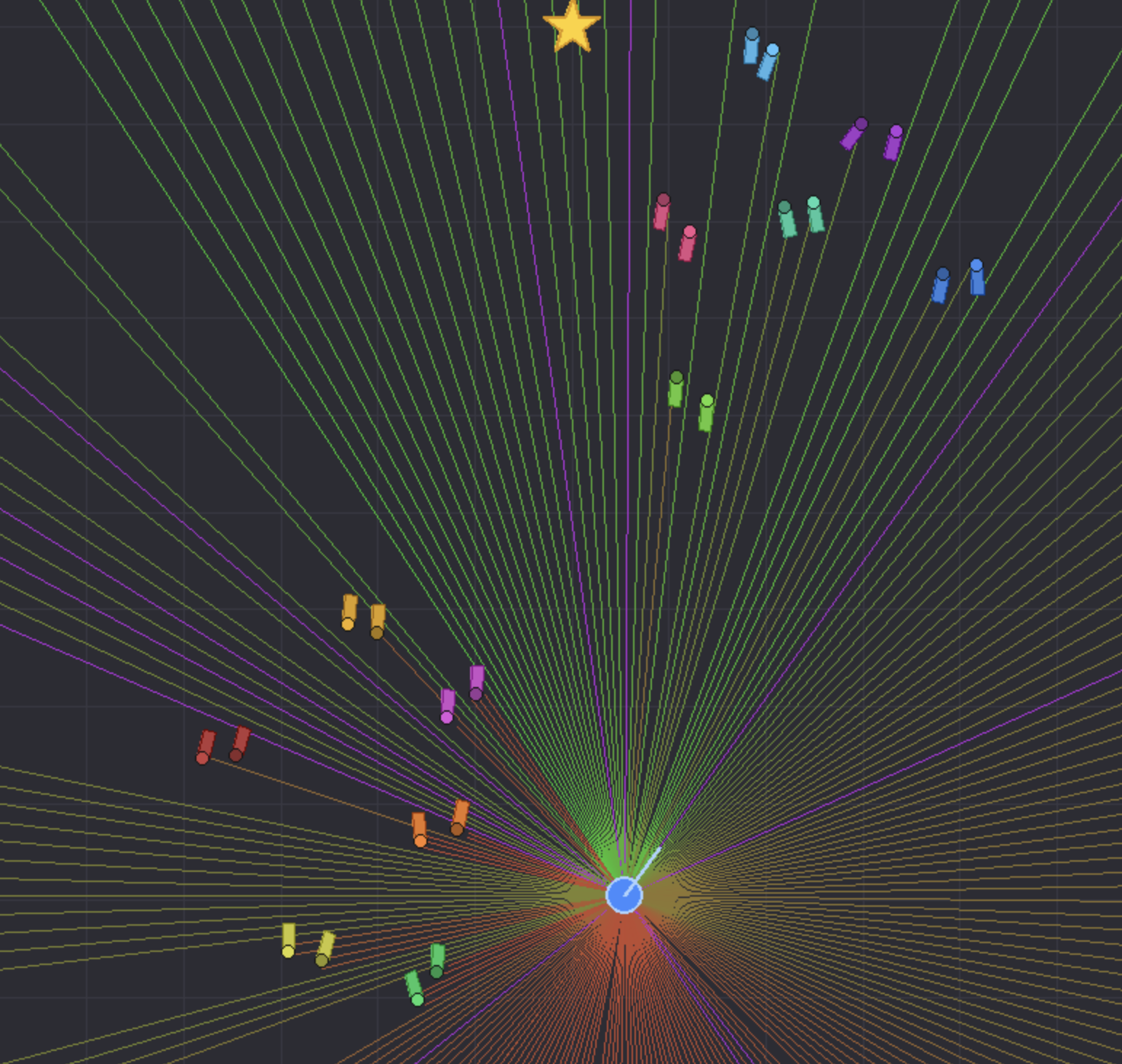}
        \caption{Zig-zag Counterflow}
    \end{subfigure}\hfill
    \begin{subfigure}[t]{0.17\linewidth}
        \includegraphics[width=\linewidth]{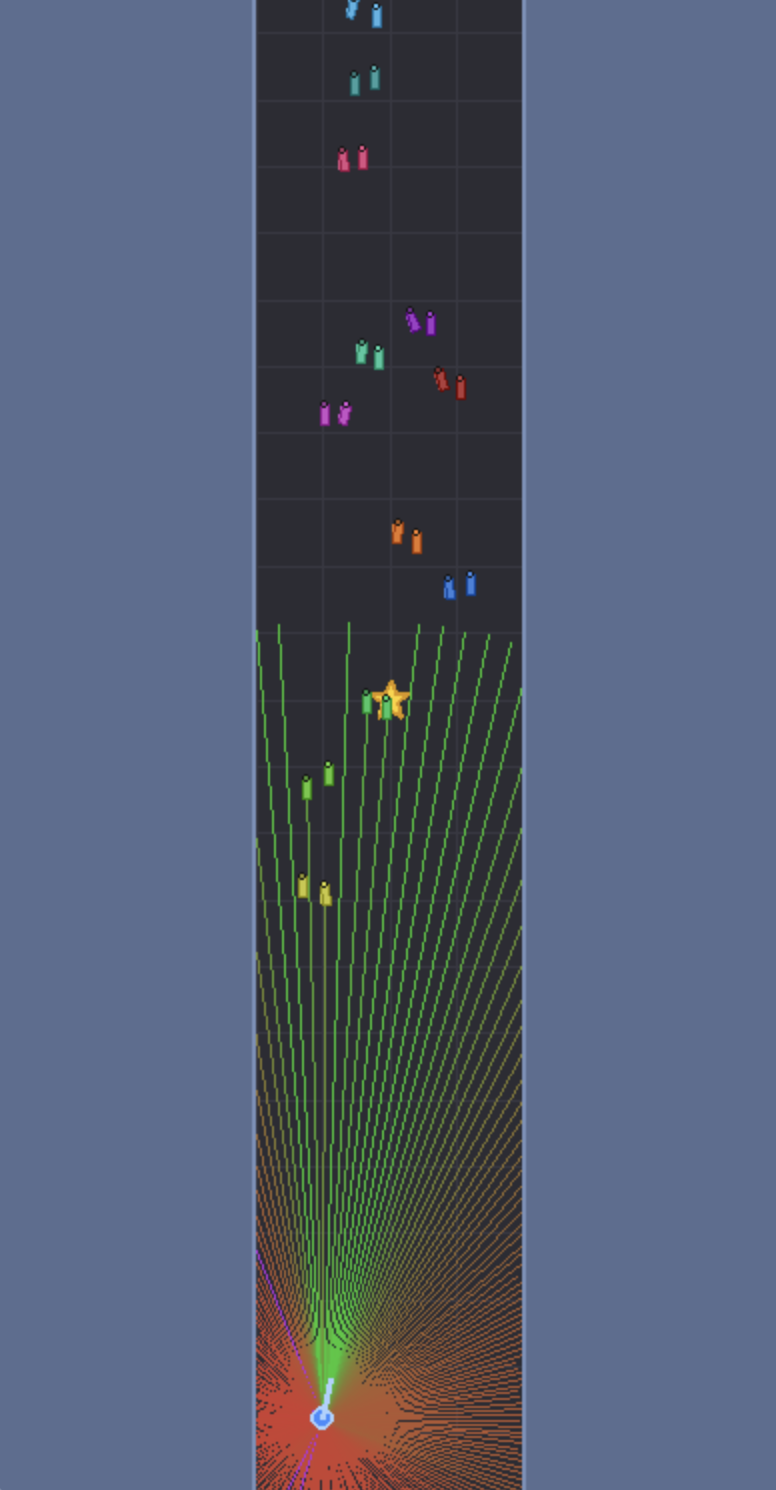}
        \caption{Long Corridor}
    \end{subfigure}\hfill
    \begin{subfigure}[t]{0.27\linewidth}
        \includegraphics[width=\linewidth]{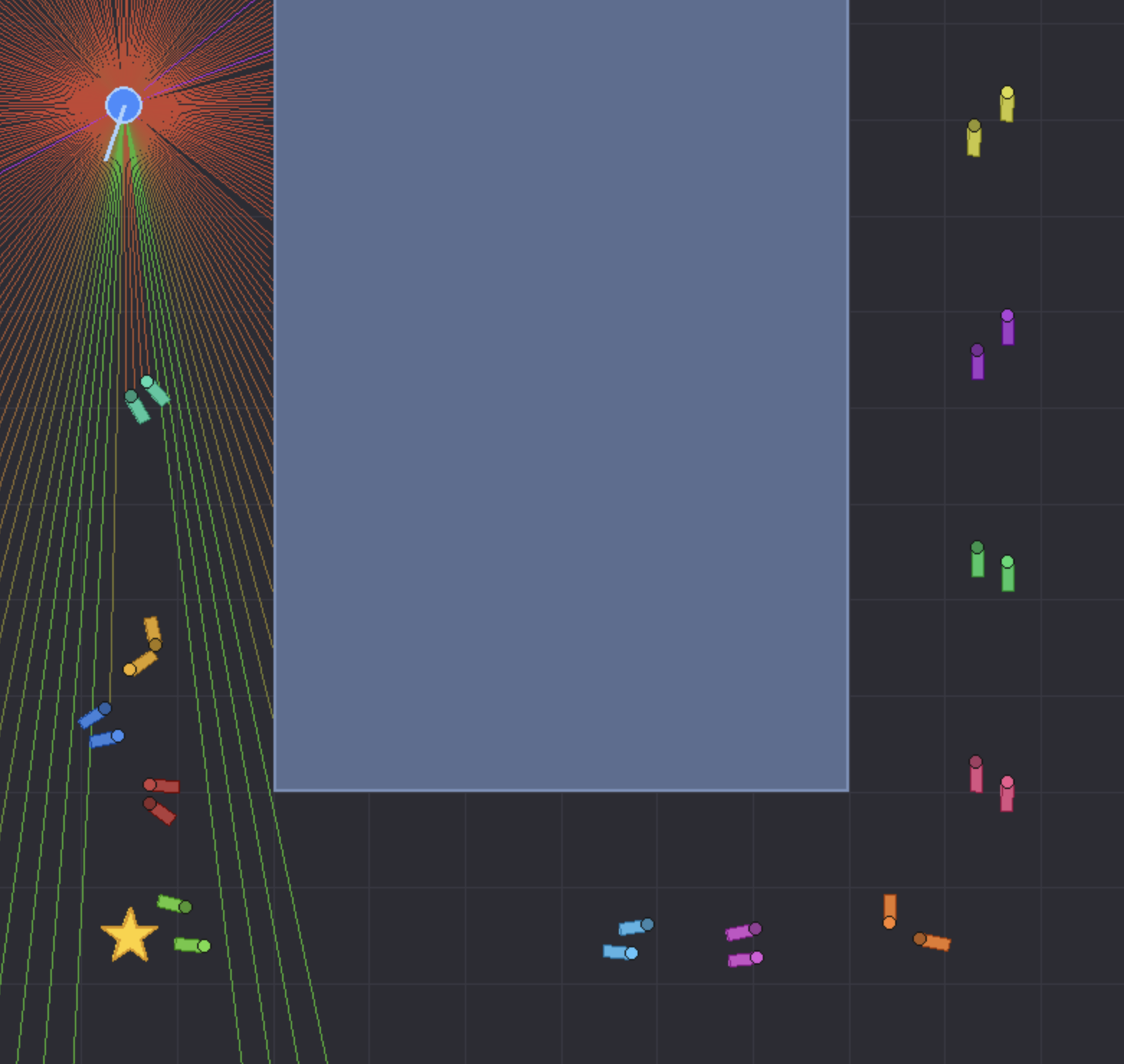}
        \caption{U-turn Corridor}
    \end{subfigure}\hfill

    \caption{Scenarios used only for testing in simulation.}
    \label{fig:testing_scenarios}
\end{figure*}

The LegNav simulator contains six additional scenarios for testing purposes only (Fig.~\ref{fig:testing_scenarios}). These are not seen during training and were designed to evaluate the generalization capabilities of the trained policies.

\begin{enumerate}
\item \textbf{S-Corridor}: s-shaped corridor with people patrolling between the walls.
\item \textbf{Sequential Rooms}: three rooms connected by two sequential bottleneck doorways with people patrolling in the rooms.
\item \textbf{Converging Crowds}: four groups of three people each moving from the four corners to the center of the room.
\item \textbf{Zig-zag Counterflow}: two groups of people walking in opposite directions with lateral shifts.
\item \textbf{Long Corridor}: similar to parallel traffic scenario but with a corridor of double length and a continuous stream of people moving in one direction.
\item \textbf{U-turn Corridor}: u-shaped corridor with people moving in the room.
\end{enumerate}

Reward weights and environment/robot parameters are summarized in Tables~\ref{tab:reward-parameters-complete}–\ref{tab:env-params}. The reward weights introduced in Sec.~\ref{subsec:reward_shaping} have been selected to balance the magnitudes of the individual reward terms.

\begin{table}[pos=!t]
\caption{Comprehensive list of reward function parameters and terminal sparse events.}
\label{tab:reward-parameters-complete}
\centering
\begin{tabular}{llc}
\toprule
\textbf{Symbol} & \textbf{Meaning} & \textbf{Value} \\
\midrule
\multicolumn{3}{l}{\textit{Dense Shaping Parameters}} \\
\midrule
$\alpha_{\text{prog}}$    & Progress reward scale                    & 1.0 \\
$\alpha_{\text{step}}$    & Step penalty / Speed reward scale        & $0.02$ \\
$\alpha_{\text{smooth}}$  & Rotational smoothness penalty scale      & $0.5$ \\
$\alpha_{\text{heading}}$ & Heading alignment reward scale           & $0.005$ \\
$\alpha_{\text{comfort}}$ & Social comfort penalty scale             & $0.15$ \\
$\alpha_{\text{yield}}$ & Yielding reward scale             & $0.1$ \\
\midrule
\multicolumn{3}{l}{\textit{Terminal Event Rewards}} \\
\midrule
$B_{\text{goal}}$            & Goal reached bonus                       & $+20.0$ \\
$B_{\text{obs}}$             & Obstacle or wall collision penalty        & $-30.0$ \\
$B_{\text{act}}$             & Active human collision penalty           & $-50.0$ \\
$B_{\text{pas}}$             & Passive human collision penalty          & $-3.5$ \\
$B_{\text{timeout}}$         & Timeout penalty                          & $-9.0$ \\
\bottomrule
\end{tabular}
\end{table}

\begin{table}[pos=!t]
\caption{Environment and robot parameters.}
\label{tab:env-params}
\centering
\begin{tabularx}{\columnwidth}{lXc}
\toprule
\textbf{Symbol} & \textbf{Description} & \textbf{Value} \\
\midrule
$L$ & LiDAR rays per scan & 216 \\
$d_{\max}$ & Max LiDAR range & $12$ m \\
$N$ & History length (stacks) & 3 \\
$\Delta t$ & Control/sim step & $0.15$ s \\
$r$ & Robot radius & $0.2$ m \\
$r_{\text{body}}$ & Pedestrian radius & $0.4$ m \\
$f_{\max}$ & Pedestrian max stepping frequency & $3.5$~$\text{s}^{-1}$ \\
$v_{\text{ref}}$ & Pedestrian reference walking speed & $2.5$~$\text{m/s}$ \\
$r_\ell$ & Radius of the pedestrian leg discs & $0.08$ m \\
$L_{\text{shoe}}$ & Pedestrian shoe length & $0.3$ m \\
$w_{\text{hip}}$ & Pedestrian hip width & $0.3$~m \\
$d_{\text{comfort}}$ & Social comfort distance threshold & $0.5\text{ m}$ \\
$d_{\text{yield}}$ & Yielding zone radius & $1.5$~m \\
$\psi_{\mathrm{FOV}}$ & Yielding zone FOV & $90^{\circ}$ \\
$v_{\max}$ & Robot max linear velocity & $\sim \mathcal{U}[0.2,\,2.0]$~m/s \\
$\omega_{\max}$ & Robot max angular velocity & $1.0$ rad/s \\
$T_{\max}$ & Max episode duration & $500$ steps \\
$v_\text{stop}$ & Speed threshold for yielding behavior & 0.1 m/s\\
\bottomrule
\end{tabularx}
\end{table}
\begin{figure}[pos=!t]
    \centering
    \includegraphics[width=\linewidth,height=0.45\textheight,keepaspectratio]{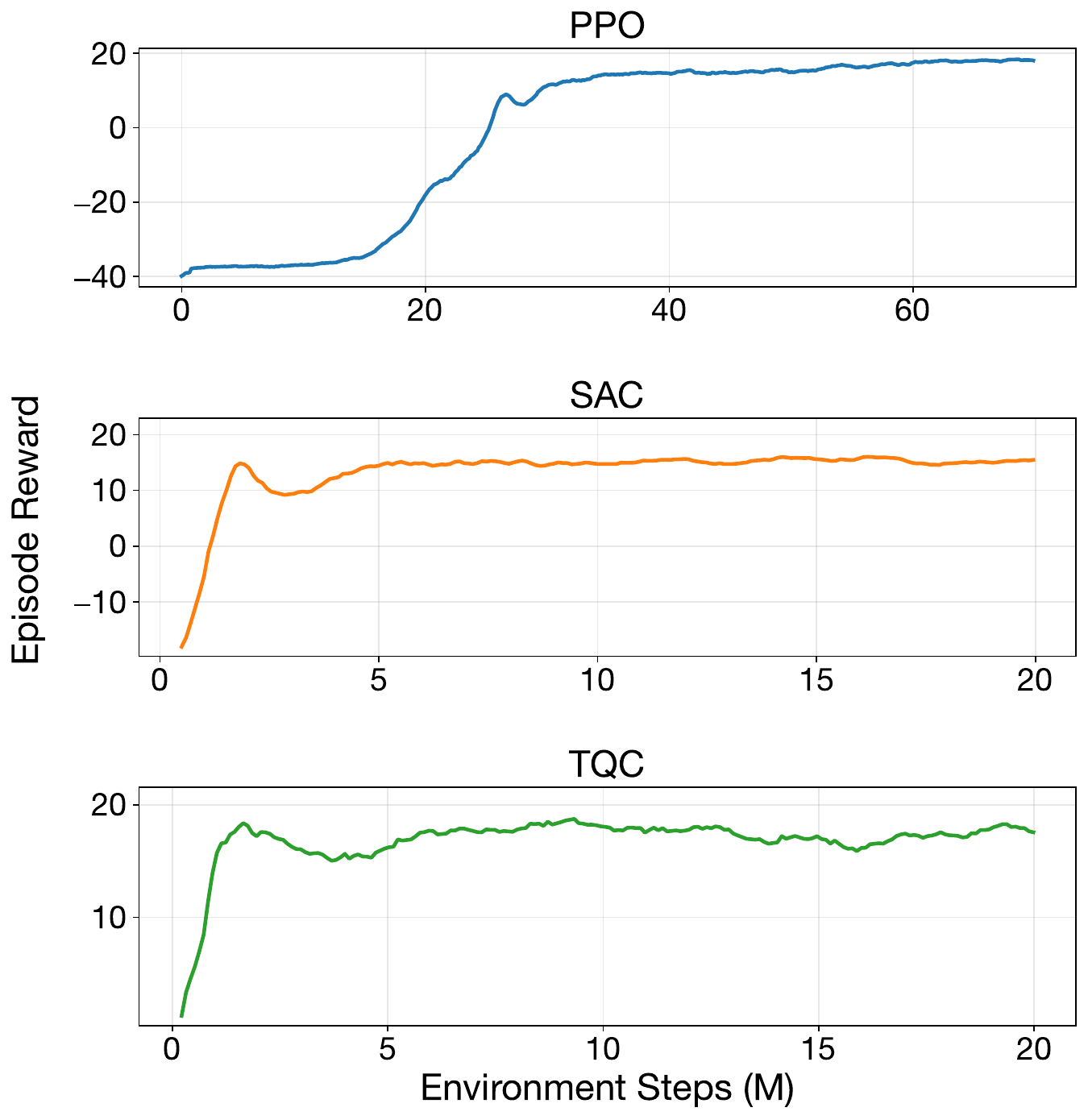}
    \caption{Episode reward (smoothed via moving average) during training for PPO, SAC and TQC.}
    \label{fig:training_rew}
\end{figure}

During training, the robot is operated in \textit{ghost mode}, meaning that pedestrians are unaware of the robot's presence. This forces the policy to learn to actively dodge pedestrians rather than relying on humans cooperatively stepping aside. During evaluation instead, ghost mode is disabled and humans react to the robot, simulating real human-robot interaction. 
All three algorithms use the same curriculum strategy, which gradually increases task difficulty as performance improves. The curriculum is driven by the agent’s rolling success rate and adjusts several aspects of training simultaneously, including goal distance, the exploration level and the set of available training scenarios.
Training begins with simpler situations and short navigation distances so the agent can first learn basic goal-seeking and collision avoidance behavior. As performance improves, more difficult scenarios and longer navigation tasks are introduced. Early scenarios focus on straightforward front-facing and lateral interactions, while more complex multi-agent environments are unlocked later in training.
To keep progression stable, the curriculum is based on a smoothed estimate of the success rate rather than instantaneous performance. The curriculum progression remains the same, even if performance temporarily drops after harder scenarios are introduced, to prevent oscillations.

We trained the CALF policy with PPO, SAC and TQC in the LegNav training scenarios. The learning curves in Fig.~\ref{fig:training_rew} reveal a stark difference in sample efficiency between the on-policy and off-policy algorithms. SAC and TQC demonstrate superior sample efficiency, with both algorithms reaching their respective plateaus around 5 million environment steps. In contrast, PPO exhibits a much slower initial learning phase, stretching out to over 50 million environment steps to reach full convergence. 
However, final performance does not perfectly correlate with sample efficiency. While SAC converges faster, it prematurely plateaus at a lower mean episode reward (around 15), reflecting its tendency toward suboptimal trajectories. Both PPO and TQC converge to a higher episode reward plateau near 18–20. PPO's learning curve is smoother throughout the training process, a direct consequence of its on-policy clipped surrogate objective, whereas the off-policy methods exhibit higher variance during their rapid initial ascent.


To overcome the CPU bottlenecks of traditional physics engines, the
entire LegNav simulator stack, including robot kinematics, pedestrian social forces,
NSG model dynamics, and LiDAR ray tracing, is implemented
entirely in JAX CUDA for GPU. The full environment step, including all force
computations, collision detection, foot updates, and sensor simulation, and the network updates,
are compiled once into a single fused GPU kernel, eliminating Python overhead between steps. This design achieves up to $135{,}000$ environment steps per second on a single RTX~3080, enabling
the learning process to converge to deployable policies in less than one hour of wall-clock time. In particular, the learning processes, depicted in Fig.~\ref{fig:training_rew}, required 41 minutes for PPO, 34 minutes for SAC, and 32 minutes for TQC.

\FloatBarrier
\section{Experimental Evaluation}
\label{sec:experiments}
Trained CALF policies are evaluated in the testing scenarios of the LegNav simulator in Fig.~\ref{fig:testing_scenarios} with different maximum speed values: $$v_{\max} \in \{0.2, 0.5, 0.75, 1.0, 1.33, 1.66, 2.0\}~\text{m/s}.$$ All policies are tested with ghost mode disabled, meaning that humans can see the robot and react to it, as they would do in reality.

The following metrics are considered for performance evaluation \cite{mavrogiannis2023core, gao2022evaluation}:\\

\noindent\textbf{Task metrics}:
\begin{itemize}
    \item \textbf{SR}: success rate, i.e., the fraction of episodes in which the robot reaches the goal within a prescribed maximum time. 
    \item \textbf{ACR}: active collision rate, i.e., fraction of episodes ending in active human collision or collisions with static obstacles.
    \item \textbf{PCR}: passive collision rate, fraction of episodes ending in passive human collision.
    \item \textbf{TR}: timeout rate, i.e., fraction of episodes ending in timeout. 
\end{itemize}
\textbf{Navigation metrics}:
\begin{itemize}
    \item \textbf{SPL}: Success-weighted by path length \cite{anderson2018evaluation}.
    \item \textbf{TTG}: average time-to-goal (s) in successful episodes.
\end{itemize}
\textbf{Social metrics}:
\begin{itemize}
    \item \textbf{MHD}: average minimum human distance (m) across the episodes. This is computed as the shortest distance between an edge of the human's shoes and the robot disc.
    \item \textbf{AJ}: average angular jerk (rad/s$^3$) in robot motion.
    \item \textbf{SC}: average space compliance, i.e., fraction of time steps in which the robot keeps a distance greater than $0.5$~m from humans.
    \item \textbf{YS} (Yielding Score): The percentage of time steps during which the robot correctly yields ($v_t \le v_\text{stop}$) when the frontal yield zone is occupied by at least one pedestrian (see (\ref{eq:yield_set})).
\end{itemize}

\subsection{Trained Algorithm Comparison}

\begin{figure*}[pos=!t]
    \centering
    \includegraphics[width=\linewidth,height=0.55\textheight,keepaspectratio]{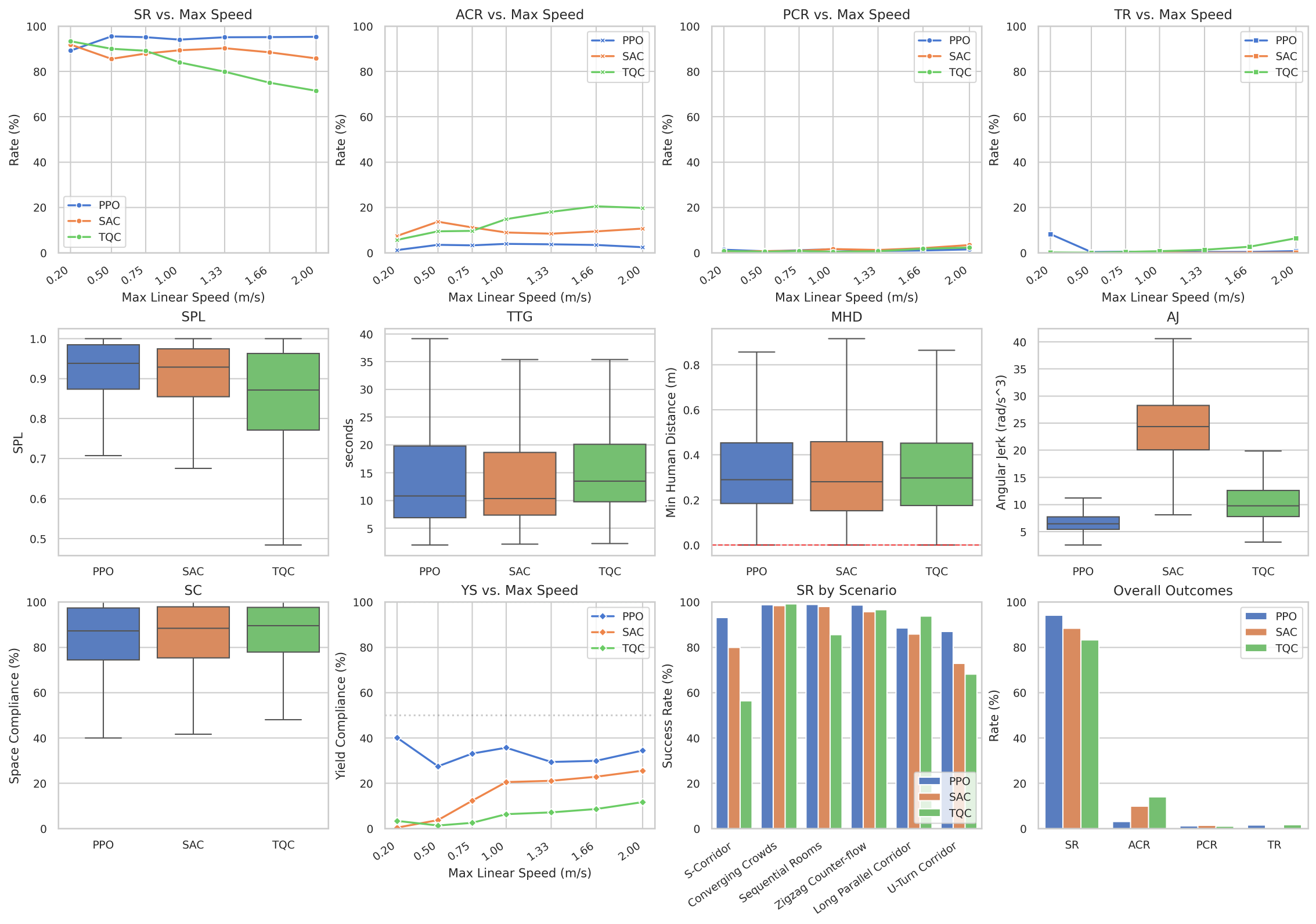}
    \caption{Evaluation dashboard comparing PPO, SAC, and TQC across speed settings, scenarios, and behavioral metrics.}
    \label{fig:dashboard}
\end{figure*}

The evaluation dashboard (Fig.~\ref{fig:dashboard}) compares the performance of the CALF policies trained with PPO, SAC, and TQC, across speed settings, behavioral metrics, and test scenarios. 

The top row reports the episode outcome rates as a function of the maximum linear speed. The success rate of PPO stays high across all speed settings. TQC instead degrades as speed grows: it starts above $90\%$ at $0.20$~m/s and falls below $80\%$ at $2.00$~m/s. The active collision rates explain this trend: TQC's collision rate climbs as speed increases, peaking near $20\%$ at $2.00$~m/s, while SAC shows a small rise at intermediate speeds and PPO stays near zero over the whole range. Passive collision rates and timeout rates are negligible at all speeds, with two minor exceptions: a small fraction of timeouts for PPO at the lowest speed ($0.20$~m/s) and for TQC at the highest ($2.00$~m/s). PPO adopts a cautious strategy and can fail to reach the goal in time when maximum speed is too low, while TQC's more aggressive policy may run out of time at higher speed, when avoidance maneuvers become harder.

The second row reports path efficiency and motion quality. PPO achieves the highest median SPL, above $0.9$, indicating more direct trajectories; TQC shows the widest variance and the lowest median SPL ($\approx 0.88$), together with the highest median time-to-goal ($\approx 13$~s). The minimum human distance distributions are similar across the three algorithms, with comparable medians, so none of the methods trades clearance for speed. The angular jerk panel separates them instead: PPO produces the smoothest trajectories, with the lowest median AJ ($\approx 6$--$7$~rad/s$^3$); SAC exhibits the highest ($\approx 25$~rad/s$^3$), indicating erratic turning; TQC falls in between.

The third row covers social compliance and generalization across environments. Space compliance stays high for all algorithms, with medians above $85\%$. Yielding Score separates the methods: PPO yields more than the others at every speed, starting near $40\%$ at $0.20$~m/s and staying around $30\%$--$35\%$; SAC and TQC start with near-zero yielding at low speeds and grow with $v_{\max}$, reaching about $25\%$ and $15\%$ at $2.00$~m/s. Across the test scenarios, PPO keeps high success rates in all layouts, while TQC struggles in the S-Corridor and Long Corridor scenarios. 
The aggregate outcomes panel summarizes the comparison: PPO obtains the highest overall success rate ($\approx 95\%$) with an active collision rate near zero; SAC follows ($\approx 88\%$); TQC ranks last ($\approx 82\%$) with the highest active collision rate ($\approx 15\%$).

\subsection{Comparison with Other Navigation Methods}

\begin{table*}[pos=!t]
\centering
\caption{Comparison of Navigation Methods — Testing Scenarios, $v_{\max} \sim \mathcal{U}[0.2,\,2.0]$~m/s. The symbols $\uparrow$ and $\downarrow$ indicate metrics respectively to be maximized and minimized.}
\label{tab:comparison}
\footnotesize
\resizebox{\textwidth}{!}{
\begin{tabular}{lccccccccccc}
\toprule
Method & Type & SR (\%) $\uparrow$ & ACR (\%) $\downarrow$ & PCR (\%) $\downarrow$ & TR (\%) $\downarrow$ & SPL $\uparrow$ & TTG (s) $\downarrow$ & MHD (m) $\uparrow$ & AJ (rad/s$^3$) $\downarrow$ & SC (\%) $\uparrow$ & YS (\%) $\uparrow$ \\
\midrule
DWA\cite{fox1997dynamic} & Model-Based & 77.7 & 0.9 & 1.2 & 20.1 & 0.97 & 22.3 & 0.30 & 1.5 & 77.3 & 64.5 \\
MPPI\cite{williams2017model} & Model-Based & 70.1 & 5.5 & 1.2 & 23.3 & 0.81 & 23.5 & 0.27 & 31.9 & 81.7 & 48.6 \\
TAGD\cite{de2024spatiotemporal} & E2E RL & 88.2 & 7.3 & 1.5 & 3.1 & 0.87 & 17.9 & 0.38 & 24.5 & 85.0 & 63.0 \\
VanillaE2E & E2E RL & 64.5 & 32.5 & 2.6 & 0.5 & 0.99 & 15.5 & 0.24 & 8.4 & 72.3 & 24.7 \\
CALF$_{PPO}$ \textit{(discs)} & E2E RL & 73.1 & 24.5 & 2.3 & 0.0 & 0.95 & 12.3 & 0.25 & 8.6 & 76.0 & 2.9 \\
CALF$_{PPO}$ & E2E RL & 95.1 & 3.3 & 0.9 & 0.7 & 0.93 & 13.7 & 0.38 & 6.8 & 83.5 & 32.4 \\
CALF$_{SAC}$ & E2E RL & 88.3 & 10.1 & 1.6 & 0.0 & 0.91 & 13.6 & 0.35 & 24.3 & 82.7 & 14.8 \\
CALF$_{TQC}$ & E2E RL & 82.8 & 14.6 & 1.0 & 1.6 & 0.84 & 15.5 & 0.36 & 11.0 & 84.9 & 5.2 \\
\bottomrule
\end{tabular}
}
\end{table*}
The three trained CALF policies, namely CALF$_{PPO}$, CALF$_{SAC}$ and CALF$_{TQC}$, are compared against alternative approaches from the navigation literature, on the testing scenarios of the LegNav simulator. Specifically, we consider classical \emph{model-based} planners, which rely on an explicit motion model: DWA~\cite{fox1997dynamic} and MPPI~\cite{williams2017model}. Furthermore, we compare against end-to-end reinforcement learning (E2E RL) paradigms that directly map raw sensor data to control commands. This last group includes TAGD~\cite{de2024spatiotemporal}, a state-of-the-art spatio-temporal RL baseline, and VanillaE2E, an ablation that replaces the CALF architecture with a plain MLP trained under the same protocol. For the CALF$_{PPO}$ policy, a variant trained with the disc approximation of pedestrians, CALF$_{PPO}$~\textit{(discs)}, is included as a comparative baseline. All methods are evaluated under the same conditions, with the per-episode maximum speed uniformly sampled in the interval $[0.2,\,2.0]$~m/s.

The results in Table~\ref{tab:comparison} show that CALF$_{PPO}$ achieves the best trade-off between task performance and social compliance. It reaches the highest success rate ($95.1\%$) and the lowest active collision rate among learning-based methods ($3.3\%$), while staying competitive on all social metrics: it ties for the largest pedestrian clearance ($0.38$~m MHD), keeps a high space compliance ($83.5\%$), and produces the second smoothest motion ($6.8$~rad/s$^3$ angular jerk, after DWA). The highest raw SPL values belong to VanillaE2E ($0.99$) and DWA ($0.97$); however, since SPL is averaged only over successful episodes, these values may reflect the efficiency of the fewer runs those policies complete, not necessarily more reliable navigation. CALF$_{PPO}$ maintains a high SPL ($0.93$) while succeeding far more often.

The classical model-based planners trade reliability for caution in dynamic environments. DWA is the most conservative method in the table: it has the lowest active collision rate ($0.9\%$), the smoothest motion ($1.5$~rad/s$^3$), and the highest yielding score ($64.5\%$). This caution, however, makes it stall: it produces the highest time-to-goal ($22.3$~s) and a high timeout rate ($20.1\%$) that caps its success at $77.7\%$. MPPI is weaker, completing only $70.1\%$ of episodes, again dominated by timeouts ($23.3\%$), with the most erratic kinematics in the table ($31.9$~rad/s$^3$) and the lowest path efficiency ($0.81$ SPL). These timeout rates reflect local-minima problems common to model-based planners, since DWA and MPPI optimize over a short horizon.

The high yielding scores of DWA and TAGD should be read together with their timeout rates and time-to-goal. The yielding score counts stopped timesteps in yield situations, so a planner that freezes near pedestrians collects yield credit without making progress. DWA's yielding score ($64.5\%$) comes with a $20.1\%$ timeout rate, which suggests that part of its apparent compliance is freezing rather than yielding. CALF$_{PPO}$ reaches a lower yielding score ($32.4\%$) but almost never times out ($0.7\%$), meaning that it stops when needed and then resumes motion.

Among the end-to-end RL baselines, TAGD is the strongest competitor, reaching $88.2\%$ success with the best space compliance ($85.0\%$) and a high yielding score ($63.0\%$). It pays for this with rougher control ($24.5$~rad/s$^3$ angular jerk, about $3.6$ times that of our policy) and more than double the active collision rate ($7.3\%$ vs.\ $3.3\%$). The VanillaE2E ablation shows the cost of removing the CALF architecture: it almost never times out ($0.5\%$) and is path-efficient on the runs it finishes, but it completes only $64.5\%$ of episodes and has the highest active collision rate of all methods ($32.5\%$). This suggests that an MLP cannot process the sensor stream safely. The off-policy CALF variants land in between: CALF$_{SAC}$ matches TAGD on success ($88.3\%$) but, like TAGD, suffers from high kinematic jerk ($24.3$~rad/s$^3$), while CALF$_{TQC}$ is smoother ($11.0$~rad/s$^3$) but more collision-prone ($14.6\%$) and almost non-yielding ($5.2\%$).

At testing time, CALF$_{PPO}$~\textit{(discs)} degrades safety: the active collision rate rises from $3.3\%$ to $24.5\%$, and the success rate falls from $95.1\%$ to $73.1\%$. The disc policy compensates with a more aggressive strategy: it achieves the shortest time-to-goal in the table ($12.3$~s) but almost never yields to pedestrians (yielding score: $2.9\%$ vs.\ $32.4\%$). The disc variant learns to pass through crowds quickly rather than navigate around them safely. This confirms that articulated leg dynamics at the sensor level are critical for learning both safe and socially compliant behavior, and are the main factor behind the benefits of the proposed method.

\begin{figure}[pos=!t]
    \centering
    \includegraphics[width=\linewidth,height=0.35\textheight,keepaspectratio]{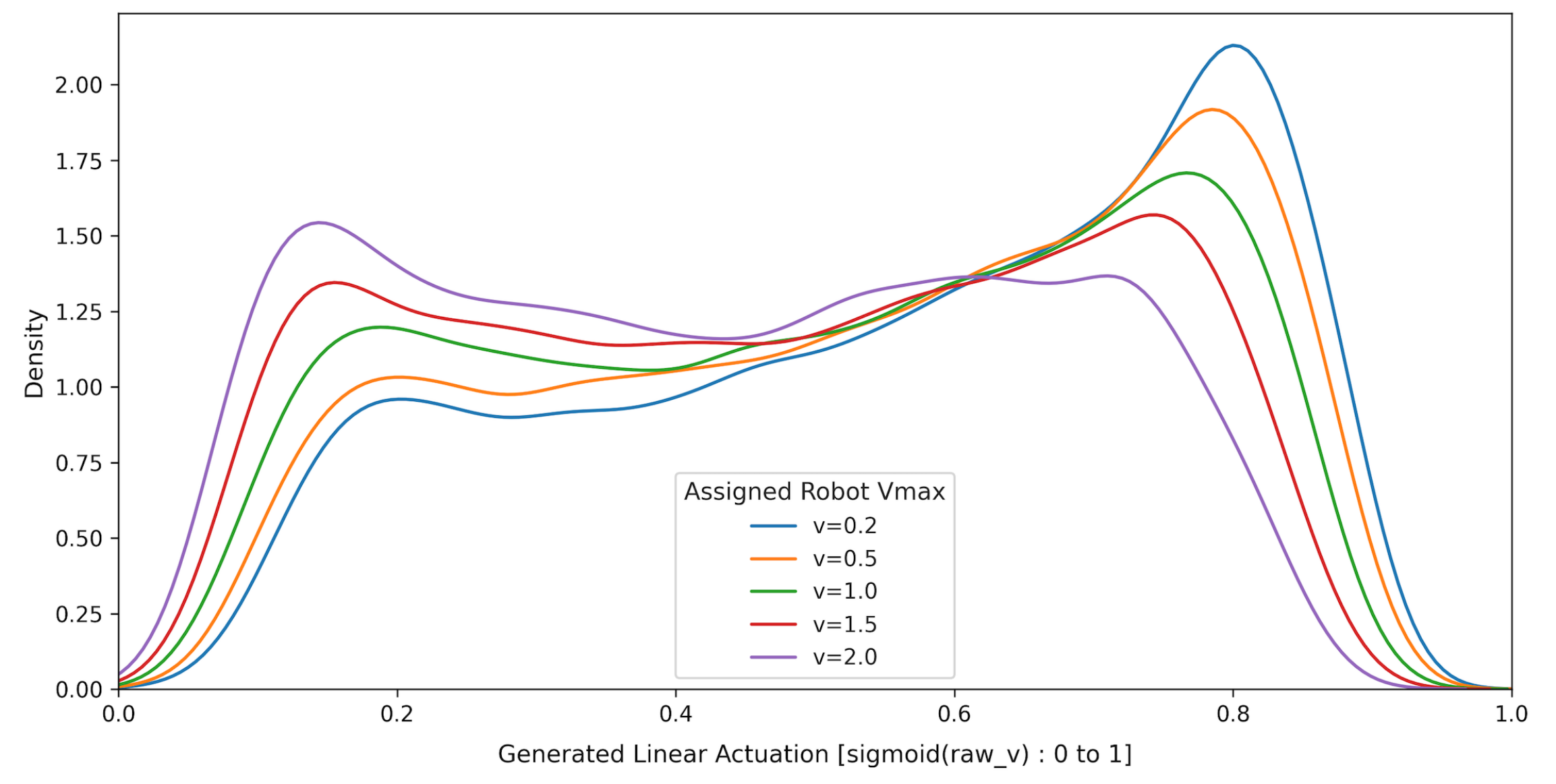}
    \caption{Distributions of commanded and effectively applied linear velocities as $v_{\max}$ varies.}
    \label{fig:param_v_max}
\end{figure}

To verify that the best trained policy CALF$_{PPO}$ correctly interprets and adapts to varying kinematic constraints, we further conducted a parametric study by modifying the maximum velocity, $v_{\max}$ (Fig.~\ref{fig:param_v_max}). The results demonstrate that the agent successfully scales its actuation based on this parameter. At lower $v_{\max}$ limits, the policy frequently operates near the maximum permitted speed to maintain efficiency. Conversely, as $v_{\max}$ increases, the velocity distribution flattens, indicating that the agent naturally learns to adopt a more conservative velocity profile to ensure safety at higher speeds.
 
\subsection{Zero-Shot Real-World Deployment}

\begin{figure}[pos=!t]
    \centering
    \includegraphics[width=0.9\linewidth,height=0.35\textheight,keepaspectratio]{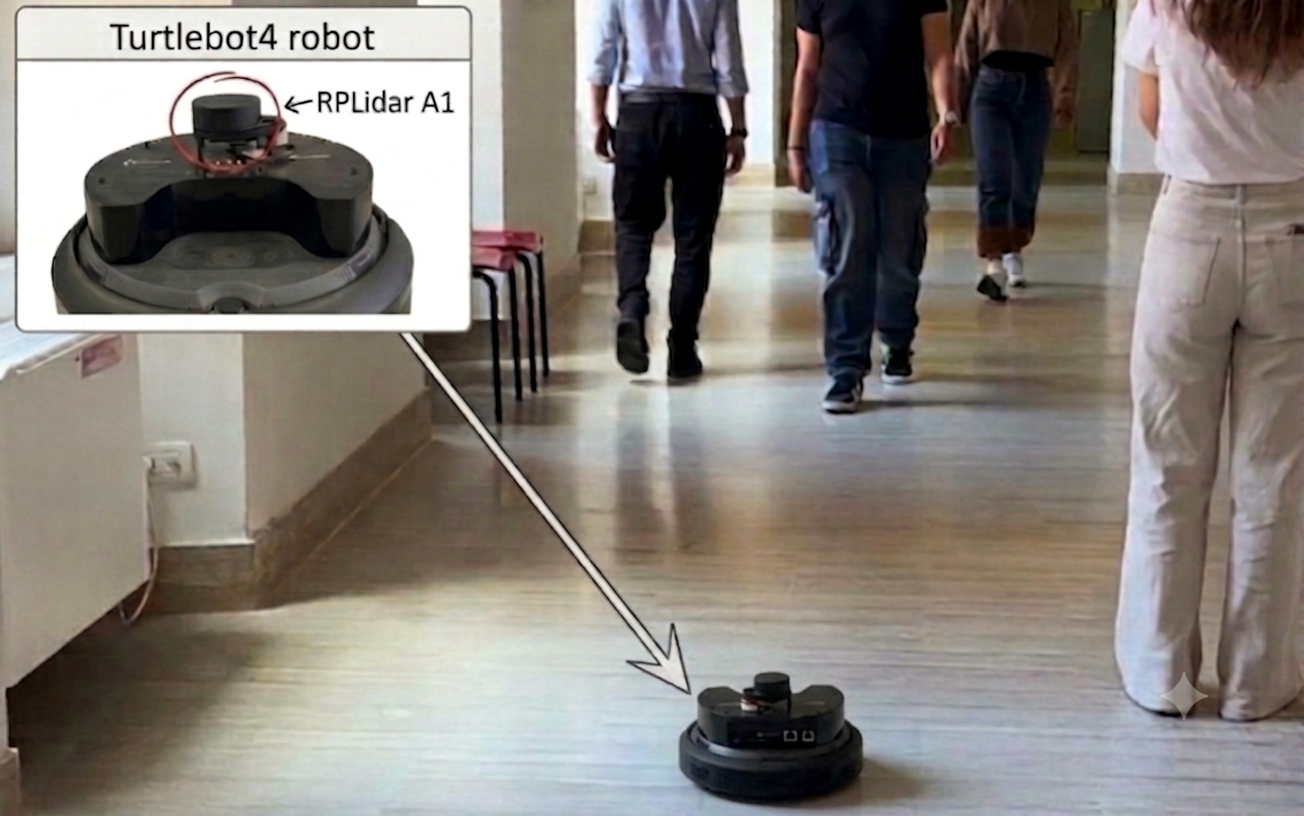}
    \caption{Real-world deployment experiments on a TurtleBot~4 platform.} 
    \label{fig:real_experiments}
\end{figure}

The CALF$_{PPO}$ policy has been zero-shot deployed on a real TurtleBot~4 platform equipped with an RPLidar A1 sensor mounted at ankle height ($\approx 0.15$~m). During experiments with 5--6 human participants walking through indoor environments (see Fig.~\ref{fig:real_experiments}), the policy proved to be successful in navigating between waypoints in several experiments. Beyond stopping, the robot demonstrated proactive social awareness, avoiding pedestrians in advance; the compliant yielding behavior was triggered only as a last resort, when the pedestrian's trajectory blocked all viable passing space.

The video of the real-world deployment is available at \url{https://www.youtube.com/watch?v=P6gFTvi3k7w}~. 

\FloatBarrier
\section{Conclusion}
\label{sec:conclusion}

This paper introduced CALF, a novel end-to-end hybrid CNN--Attention--MLP architecture, specifically designed to process the alternating leg LiDAR signatures of pedestrians for social navigation. 
To supply this architecture with realistic sensory inputs, we introduced the NSG model, which simulates human gait in 2D. It reproduces the characteristic two-cluster LiDAR signature each pedestrian generates at ankle height, capturing the articulated motion of the legs rather than approximating pedestrians as discs. Our disc-to-leg comparative analysis isolates the effect: replacing NSG with the standard disc model raised the active collision rate and collapsed yielding compliance. This confirms that simulating articulated leg dynamics at the sensor level is a key factor for reducing the sim-to-real gap and thus learning safe and effective navigation policies. 

Benchmarking against model-based and end-to-end RL methods on unseen, dynamic scenarios showed that the CALF policy, trained with PPO, achieves an optimal balance of task performance, safety, and social compliance. It reaches the highest success rate with the lowest active collision rate among learning-based methods, while maintaining smooth motion and competitive social metrics. Furthermore, our comparisons expose a dilemma common to existing approaches. Classical model-based planners are safe but over-conservative, stalling in local minima instead of committing to a path, and generating apparent social compliance by freezing rather than yielding. Learning-based baselines, in contrast, tend to privilege throughput; yet, they do it by giving up either safety guarantees or motion smoothness. The proposed CALF policy achieves a good trade-off between social compliance and navigation efficiency.

The zero-shot deployment of the CALF policy on a TurtleBot~4, achieved with no fine-tuning or domain adaptation, provided a real-world validation of the simulation results.

Future work will focus on improving latent feature representation within the end-to-end network, for instance, by introducing auxiliary pedestrian-detection losses during training. This guides the policy to better recognize real human motion without requiring a decoupled detection pipeline at deployment.




\bibliographystyle{elsarticle-num}
\bibliography{legnav}

@INPROCEEDINGS{arras2007using,
  author={Arras, Kai O. and Mozos, Oscar Martinez and Burgard, Wolfram},
  booktitle={Proceedings 2007 IEEE International Conference on Robotics and Automation}, 
  title={Using Boosted Features for the Detection of People in 2D Range Data}, 
  year={2007},
  volume={},
  number={},
  pages={3402-3407},
  doi={10.1109/ROBOT.2007.363998}}

@INPROCEEDINGS{long2018towards,
  author={Long, Pinxin and Fan, Tingxiang and Liao, Xinyi and Liu, Wenxi and Zhang, Hao and Pan, Jia},
  booktitle={2018 IEEE International Conference on Robotics and Automation (ICRA)}, 
  title={Towards Optimally Decentralized Multi-Robot Collision Avoidance via Deep Reinforcement Learning}, 
  year={2018},
  volume={},
  number={},
  pages={6252-6259},
  doi={10.1109/ICRA.2018.8461113}}

@misc{schulman2017proximal,
  author  = {Schulman, John and Wolski, Filip and Dhariwal, Prafulla and Radford, Alec and Klimov, Oleg},
  title   = {Proximal Policy Optimization Algorithms},
  year    = {2017},
  eprint  = {1707.06347},
  archivePrefix = {arXiv},
  primaryClass  = {cs.LG},
  url     = {https://arxiv.org/abs/1707.06347}
}

@inproceedings{haarnoja2018soft,
  author    = {Haarnoja, Tuomas and Zhou, Aurick and Abbeel, Pieter and Levine, Sergey},
  title     = {Soft Actor-Critic: Off-Policy Maximum Entropy Deep Reinforcement Learning with a Stochastic Actor},
  booktitle = {Proceedings of the 35th International Conference on Machine Learning (ICML)},
  year      = {2018},
  pages     = {1861--1870}
}

@InProceedings{kuznetsov2020controlling,
  title = 	 {Controlling Overestimation Bias with Truncated Mixture of Continuous Distributional Quantile Critics},
  author =       {Kuznetsov, Arsenii and Shvechikov, Pavel and Grishin, Alexander and Vetrov, Dmitry},
  booktitle = 	 {Proceedings of the 37th International Conference on Machine Learning},
  pages = 	 {5556--5566},
  year = 	 {2020},
  editor = 	 {III, Hal Daumé and Singh, Aarti},
  volume = 	 {119},
  series = 	 {Proceedings of Machine Learning Research},
  month = 	 {13--18 Jul},
  publisher =    {PMLR}
}

@article{helbing1995social,
  author  = {Helbing, Dirk and Moln{\'a}r, P{\'e}ter},
  title   = {Social force model for pedestrian dynamics},
  journal = {Physical Review E},
  volume  = {51},
  number  = {5},
  pages   = {4282--4286},
  year    = {1995},
  publisher = {American Physical Society}
}

@article{farina2017walking,
  author  = {Farina, Francesco and Fontanelli, Daniele and Garulli, Andrea and Giannitrapani, Antonio and Prattichizzo, Domenico},
  title   = {Walking Ahead: The Headed Social Force Model},
  journal = {PLOS ONE},
  volume  = {12},
  number  = {1},
  pages   = {e0169734},
  year    = {2017},
  publisher = {Public Library of Science}
}

@INPROCEEDINGS{leigh2015person,
  author={Leigh, Angus and Pineau, Joelle and Olmedo, Nicolas and Zhang, Hong},
  booktitle={2015 IEEE International Conference on Robotics and Automation (ICRA)}, 
  title={Person tracking and following with 2D laser scanners}, 
  year={2015},
  volume={},
  number={},
  pages={726-733},
  doi={10.1109/ICRA.2015.7139259}}

@InProceedings{van2011reciprocal,
author="van den Berg, Jur
and Guy, Stephen J.
and Lin, Ming
and Manocha, Dinesh",
editor="Pradalier, C{\'e}dric
and Siegwart, Roland
and Hirzinger, Gerhard",
title="Reciprocal n-Body Collision Avoidance",
booktitle="Robotics Research",
year="2011",
publisher="Springer Berlin Heidelberg",
address="Berlin, Heidelberg",
pages="3--19",
isbn="978-3-642-19457-3"
}

@INPROCEEDINGS{wang2023navistar,
  author={Wang, Weizheng and Wang, Ruiqi and Mao, Le and Min, Byung-Cheol},
  booktitle={2023 IEEE/RSJ International Conference on Intelligent Robots and Systems (IROS)}, 
  title={NaviSTAR: Socially Aware Robot Navigation with Hybrid Spatio-Temporal Graph Transformer and Preference Learning}, 
  year={2023},
  volume={},
  number={},
  pages={11348-11355},
  doi={10.1109/IROS55552.2023.10341395}}

@ARTICLE{fox1997dynamic,
  author={Fox, D. and Burgard, W. and Thrun, S.},
  journal={IEEE Robotics \& Automation Magazine}, 
  title={The dynamic window approach to collision avoidance}, 
  year={1997},
  volume={4},
  number={1},
  pages={23-33},
  doi={10.1109/100.580977}}

@INPROCEEDINGS{chen2017decentralized,
  author={Chen, Yu Fan and Liu, Miao and Everett, Michael and How, Jonathan P.},
  booktitle={2017 IEEE International Conference on Robotics and Automation (ICRA)}, 
  title={Decentralized non-communicating multiagent collision avoidance with deep reinforcement learning}, 
  year={2017},
  volume={},
  number={},
  pages={285-292},
  doi={10.1109/ICRA.2017.7989037}}

@INPROCEEDINGS{chen2017socially,
  author={Chen, Yu Fan and Everett, Michael and Liu, Miao and How, Jonathan P.},
  booktitle={2017 IEEE/RSJ International Conference on Intelligent Robots and Systems (IROS)}, 
  title={Socially aware motion planning with deep reinforcement learning}, 
  year={2017},
  volume={},
  number={},
  pages={1343-1350},
  doi={10.1109/IROS.2017.8202312}}

@INPROCEEDINGS{chen2019crowd,
  author={Chen, Changan and Liu, Yuejiang and Kreiss, Sven and Alahi, Alexandre},
  booktitle={2019 International Conference on Robotics and Automation (ICRA)}, 
  title={Crowd-Robot Interaction: Crowd-Aware Robot Navigation With Attention-Based Deep Reinforcement Learning}, 
  year={2019},
  volume={},
  number={},
  pages={6015-6022},
  doi={10.1109/ICRA.2019.8794134}}

@ARTICLE{xie2023drl,
  author={Xie, Zhanteng and Dames, Philip},
  journal={IEEE Transactions on Robotics}, 
  title={DRL-VO: Learning to Navigate Through Crowded Dynamic Scenes Using Velocity Obstacles}, 
  year={2023},
  volume={39},
  number={4},
  pages={2700-2719},
  doi={10.1109/TRO.2023.3257549}}

@INPROCEEDINGS{zhu2024learn,
  author={Zhu, Wei and Hayashibe, Mitsuhiro},
  booktitle={2024 IEEE International Conference on Robotics and Automation (ICRA)}, 
  title={Learn to Navigate in Dynamic Environments with Normalized LiDAR Scans}, 
  year={2024},
  volume={},
  number={},
  pages={7568-7575},
  doi={10.1109/ICRA57147.2024.10611247}}

@article{martinez2025rumor,
title = {RUMOR: Reinforcement learning for understanding a model of the real world for navigation in dynamic environments},
journal = {Robotics and Autonomous Systems},
volume = {191},
pages = {105020},
year = {2025},
issn = {0921-8890},
doi = {https://doi.org/10.1016/j.robot.2025.105020},
url = {https://www.sciencedirect.com/science/article/pii/S092188902500106X},
author = {Diego Martinez-Baselga and Luis Riazuelo and Luis Montano}
}

@ARTICLE{gao2022evaluation,  
AUTHOR={Gao, Yuxiang  and Huang, Chien-Ming },        
TITLE={Evaluation of Socially-Aware Robot Navigation},     
JOURNAL={Frontiers in Robotics and AI},        
VOLUME={Volume 8 - 2021},
YEAR={2022},
DOI={10.3389/frobt.2021.721317},
ISSN={2296-9144}
}

@article{mavrogiannis2023core,
author = {Mavrogiannis, Christoforos and Baldini, Francesca and Wang, Allan and Zhao, Dapeng and Trautman, Pete and Steinfeld, Aaron and Oh, Jean},
title = {Core Challenges of Social Robot Navigation: A Survey},
year = {2023},
issue_date = {September 2023},
publisher = {Association for Computing Machinery},
address = {New York, NY, USA},
volume = {12},
number = {3},
url = {https://doi.org/10.1145/3583741},
doi = {10.1145/3583741},
journal = {J. Hum.-Robot Interact.},
month = apr,
articleno = {36},
numpages = {39}
}

@ARTICLE{yang2023st,
  author={Yang, Yuxiang and Jiang, Jiahao and Zhang, Jing and Huang, Jiye and Gao, Mingyu},
  journal={IEEE Robotics and Automation Letters}, 
  title={ST$^{2}$: Spatial-Temporal State Transformer for Crowd-Aware Autonomous Navigation}, 
  year={2023},
  volume={8},
  number={2},
  pages={912-919},
  doi={10.1109/LRA.2023.3234815}}

@INPROCEEDINGS{dugas2021navrep,
  author={Dugas, Daniel and Nieto, Juan and Siegwart, Roland and Chung, Jen Jen},
  booktitle={2021 IEEE International Conference on Robotics and Automation (ICRA)}, 
  title={NavRep: Unsupervised Representations for Reinforcement Learning of Robot Navigation in Dynamic Human Environments}, 
  year={2021},
  volume={},
  number={},
  pages={7829-7835},
  doi={10.1109/ICRA48506.2021.9560951}
}

@INPROCEEDINGS{everett2018motion,
  author={Everett, Michael and Chen, Yu Fan and How, Jonathan P.},
  booktitle={2018 IEEE/RSJ International Conference on Intelligent Robots and Systems (IROS)}, 
  title={Motion Planning Among Dynamic, Decision-Making Agents with Deep Reinforcement Learning}, 
  year={2018},
  volume={},
  number={},
  pages={3052-3059},
  doi={10.1109/IROS.2018.8593871}}

@ARTICLE{everett2021collision,
  author={Everett, Michael and Chen, Yu Fan and How, Jonathan P.},
  journal={IEEE Access}, 
  title={Collision Avoidance in Pedestrian-Rich Environments With Deep Reinforcement Learning}, 
  year={2021},
  volume={9},
  number={},
  pages={10357-10377},
  doi={10.1109/ACCESS.2021.3050338}}

@article{williams2017model,
  title={Model predictive path integral control: From theory to parallel computation},
  author={Williams, Grady and Aldrich, Andrew and Theodorou, Evangelos A},
  journal={Journal of Guidance, Control, and Dynamics},
  volume={40},
  number={2},
  pages={344--357},
  year={2017},
  publisher={American Institute of Aeronautics and Astronautics}
}

@INPROCEEDINGS{perez2021robot,
  author={Pérez-D’Arpino, Claudia and Liu, Can and Goebel, Patrick and Martín-Martín, Roberto and Savarese, Silvio},
  booktitle={2021 IEEE International Conference on Robotics and Automation (ICRA)}, 
  title={Robot Navigation in Constrained Pedestrian Environments using Reinforcement Learning}, 
  year={2021},
  volume={},
  number={},
  pages={1140-1146},
  doi={10.1109/ICRA48506.2021.9560893}}

@INPROCEEDINGS{jin2020mapless,
  author={Jin, Jun and Nguyen, Nhat M. and Sakib, Nazmus and Graves, Daniel and Yao, Hengshuai and Jagersand, Martin},
  booktitle={2020 IEEE International Conference on Robotics and Automation (ICRA)}, 
  title={Mapless Navigation among Dynamics with Social-safety-awareness: a reinforcement learning approach from 2D laser scans}, 
  year={2020},
  volume={},
  number={},
  pages={6979-6985},
  doi={10.1109/ICRA40945.2020.9197148}}

@ARTICLE{de2024spatiotemporal,
  author={de Heuvel, Jorge and Zeng, Xiangyu and Shi, Weixian and Sethuraman, Tharun and Bennewitz, Maren},
  journal={IEEE Robotics and Automation Letters}, 
  title={Spatiotemporal Attention Enhances Lidar-Based Robot Navigation in Dynamic Environments}, 
  year={2024},
  volume={9},
  number={5},
  pages={4202-4209},
  doi={10.1109/LRA.2024.3373988}}

@INPROCEEDINGS{chen2020relational,
  author={Chen, Changan and Hu, Sha and Nikdel, Payam and Mori, Greg and Savva, Manolis},
  booktitle={2020 IEEE/RSJ International Conference on Intelligent Robots and Systems (IROS)}, 
  title={Relational Graph Learning for Crowd Navigation}, 
  year={2020},
  volume={},
  number={},
  pages={10007-10013},
  doi={10.1109/IROS45743.2020.9340705}}

@article{dabney2018distributional, 
    title={Distributional Reinforcement Learning With Quantile Regression}, 
    volume={32}, 
    DOI={10.1609/aaai.v32i1.11791}, 
    abstractNote={ &amp;lt;p&amp;gt; In reinforcement learning (RL), an agent interacts with the environment by taking actions and observing the next state and reward. When sampled probabilistically, these state transitions, rewards, and actions can all induce randomness in the observed long-term return. Traditionally, reinforcement learning algorithms average over this randomness to estimate the value function. In this paper, we build on recent work advocating a distributional approach to reinforcement learning in which the distribution over returns is modeled explicitly instead of only estimating the mean. That is, we examine methods of learning the value distribution instead of the value function. We give results that close a number of gaps between the theoretical and algorithmic results given by Bellemare, Dabney, and Munos (2017). First, we extend existing results to the approximate distribution setting. Second, we present a novel distributional reinforcement learning algorithm consistent with our theoretical formulation. Finally, we evaluate this new algorithm on the Atari 2600 games, observing that it significantly outperforms many of the recent improvements on DQN, including the related distributional algorithm C51. &amp;lt;/p&amp;gt; }, 
    number={1}, 
    journal={Proceedings of the AAAI Conference on Artificial Intelligence}, author={Dabney, Will and Rowland, Mark and Bellemare, Marc and Munos, Rémi}, 
    year={2018}, 
    month={Apr.} 
}

@INPROCEEDINGS{bae2026deep,
  author={Bae, Sang Uk and Han, Dong Seog},
  booktitle={2026 International Conference on Artificial Intelligence in Information and Communication (ICAIIC)}, 
  title={Deep Reinforcement Learning-Based Mobile Robot Navigation Using Truncated Quantile Critics}, 
  year={2026},
  volume={},
  number={},
  pages={1369-1401},
  doi={10.1109/ICAIIC68212.2026.11454248}
}

@article{choton2025efficient,
  title={Efficient environment design for multi-robot navigation via continuous control},
  author={Choton, Jahid Chowdhury and Woods, John and Hsu, William},
  journal={arXiv preprint arXiv:2508.14105},
  year={2025}
}

@inproceedings{bellemare2017distributional,
  author    = {Bellemare, Marc G. and Dabney, Will and Munos, R{\'e}mi},
  title     = {A distributional perspective on reinforcement learning},
  booktitle = {Proceedings of the 34th International Conference on Machine Learning (ICML)},
  year      = {2017},
  pages     = {449--458}
}

@inproceedings{chen2021robotnav,
  author    = {Chen, Xi and Everett, Michael and Liu, Mo and How, Jonathan P.},
  title     = {Robot Navigation in Crowds by Implicit Cooperative Modeling},
  booktitle = {Proceedings of the AAAI Conference on Artificial Intelligence},
  volume    = {35},
  pages     = {11303--11311},
  year      = {2021},
  doi       = {10.1609/aaai.v35i13.17258}
}

@inproceedings{rajeswaran2017towards,
  author    = {Rajeswaran, Aravind and Lowrey, Kendall and Todorov, Emanuel and Kakade, Sham},
  title     = {Towards generalization and simplicity in continuous control},
  booktitle = {Proceedings of the 31st International Conference on Neural Information Processing Systems (NeurIPS)},
  pages     = {6553--6564},
  year      = {2017}
}

@misc{jax2018github,
  author  = {James Bradbury and Roy Frostig and Peter Hawkins and
             Matthew James Johnson and Chris Leary and Dougal Maclaurin and
             George Necula and Adam Paszke and Jake Vander{P}las and
             Skye Wanderman-{M}ilne and Qiao Zhang},
  title   = {{JAX}: composable transformations of {P}ython+{N}um{P}y programs},
  year    = {2018},
  note    = {Version 0.3.13. Available at \url{http://github.com/jax-ml/jax}},
}

@article{haarnoja2019learning,
  author  = {Haarnoja, Tuomas and Zhou, Aurick and Abbeel, Pieter and Levine, Sergey},
  title   = {Learning to Walk via Deep Reinforcement Learning},
  journal = {Proceedings of Robotics: Science and Systems (RSS)},
  year    = {2019},
  doi     = {10.15607/RSS.2019.XV.011}
}

@inproceedings{ma2022sacnavigation,
  author    = {Ma, Yiming and Liu, Yuhang and Wang, Jian and Yu, Wenjing},
  title     = {Socially Compliant Navigation with Soft Actor-Critic Reinforcement Learning},
  booktitle = {Proceedings of the IEEE International Conference on Mechatronics and Automation (ICMA)},
  year      = {2022},
  pages     = {1734--1739},
  doi       = {10.1109/ICMA54519.2022.9914739}
}

@inproceedings{henderson2018deep,
  author    = {Henderson, Peter and Islam, Riashat and Bachman, Philip and
               Pineau, Joelle and Precup, Doina and Meger, David},
  title     = {Deep Reinforcement Learning That Matters},
  booktitle = {Proceedings of the Thirty-Second AAAI Conference on Artificial Intelligence},
  pages     = {3207--3214},
  year      = {2018},
  doi       = {10.1609/aaai.v32i1.11694}
}

@inproceedings{haarnoja2018applications,
  author    = {Haarnoja, Tuomas and Zhou, Aurick and Hartikainen, Kristian and Tucker, George and Ha, Sehoon and Tan, Jie and Kumar, Vikash and Zhu, Henry and Gupta, Abhishek and Abbeel, Pieter and Levine, Sergey},
  title     = {Soft Actor-Critic Algorithms and Applications},
  booktitle = {Proceedings of the 2nd Conference on Robot Learning (CoRL)},
  year      = {2018},
  pages     = {1--20},
  url       = {https://proceedings.mlr.press/v87/haarnoja18b.html}
}

@inproceedings{konda1999actor,
  title     = {Actor-Critic Algorithms},
  author    = {Konda, Vijay R. and Tsitsiklis, John N.},
  booktitle = {Advances in Neural Information Processing Systems 12},
  pages     = {1008--1014},
  year      = {1999}
}

@book{thrun2005probabilistic,
  title={Probabilistic Robotics},
  author={Thrun, Sebastian and Burgard, Wolfram and Fox, Dieter},
  year={2005},
  publisher={MIT Press}
}

@inproceedings{isele2018navigating,
  author    = {Isele, David and Cosgun, Akansel and Subramanian, Kaustubh and Booth, Kyle and Christensen, Henrik I.},
  title     = {Navigating occluded intersections with autonomous vehicles using deep reinforcement learning},
  booktitle = {Proceedings of the IEEE International Conference on Robotics and Automation (ICRA)},
  pages     = {2034--2039},
  year      = {2018},
  doi       = {10.1109/ICRA.2018.8460766}
}

@inproceedings{zou2020reducing,
  title={Reducing footskate in human motion reconstruction with ground contact constraints},
  author={Zou, Yuliang and Yang, Jimei and Ceylan, Duygu and Zhang, Jianming and Perazzi, Federico and Huang, Jia-Bin},
  booktitle={Proceedings of the IEEE/CVF Winter Conference on Applications of Computer Vision},
  pages={459--468},
  year={2020}
}

@article{boulic1990global,
  title={A global human walking model with real-time kinematic personification},
  author={Boulic, Ronan and Thalmann, Nadia Magnenat and Thalmann, Daniel},
  journal={The visual computer},
  volume={6},
  number={6},
  pages={344--358},
  year={1990},
  publisher={Springer}
}

@article{beacco2015footstep,
  title={Footstep parameterized motion blending using barycentric coordinates},
  author={Beacco, Alejandro and Pelechano, Nuria and Kapadia, Mubbasir and Badler, Norman I},
  journal={Computers \& Graphics},
  volume={47},
  pages={105--112},
  year={2015},
  publisher={Elsevier}
}

@article{fukuchi2019effects,
  title={Effects of walking speed on gait biomechanics in healthy participants: a systematic review and meta-analysis},
  author={Fukuchi, Claudiane Arakaki and Fukuchi, Reginaldo Kisho and Duarte, Marcos},
  journal={Systematic reviews},
  volume={8},
  number={1},
  pages={153},
  year={2019},
  publisher={Springer}
}

@article{anderson2018evaluation,
  title={On evaluation of embodied navigation agents},
  author={Anderson, Peter and Chang, Angel and Chaplot, Devendra Singh and Dosovitskiy, Alexey and Gupta, Saurabh and Koltun, Vladlen and Kosecka, Jana and Malik, Jitendra and Mottaghi, Roozbeh and Savva, Manolis and others},
  journal={arXiv preprint arXiv:1807.06757},
  year={2018}
}

@ARTICLE{stratton2025characterizing,
  author={Stratton, Andrew and Hauser, Kris and Mavrogiannis, Christoforos},
  journal={IEEE Robotics and Automation Letters}, 
  title={Characterizing the Complexity of Social Robot Navigation Scenarios}, 
  year={2025},
  volume={10},
  number={1},
  pages={184-191},
  doi={10.1109/LRA.2024.3502060}}

@ARTICLE{ahsen2025domain,
  author={Sen, Nick Ah and Kulić, Dana and Carreno-Medrano, Pamela},
  journal={IEEE Robotics and Automation Letters}, 
  title={Domain Randomization for Learning to Navigate in Human Environments}, 
  year={2025},
  volume={10},
  number={2},
  pages={1625-1632},
  doi={10.1109/LRA.2024.3521178}}

\end{document}